%% file: example_paper.tex
\documentclass{article}
\usepackage{graphicx}  
\usepackage{xcolor}   
\usepackage{makecell}  
\usepackage{microtype}
\usepackage{graphicx}
\usepackage{subcaption}
\usepackage{booktabs} % for professional tables

\usepackage{hyperref}

\usepackage{threeparttable}
\usepackage{tcolorbox}
\usepackage{mdframed,xcolor}
\usepackage[preprint]{icml2026}
\usepackage[table]{xcolor} 
\usepackage{amsmath}
\usepackage{amssymb}
\usepackage{mathtools}
\usepackage{amsthm}
\usepackage{adjustbox}

\usepackage{booktabs}
\usepackage{multirow}
\usepackage[capitalize,noabbrev]{cleveref}
\definecolor{LightPurple}{HTML}{E6E6FA} 
\definecolor{lightblue}{RGB}{235, 244, 246}
\newcommand{\gainG}[1]{\textcolor{blue}{\footnotesize\textit{(#1)}}}   % vs Gemini
\theoremstyle{plain}
\newtheorem{theorem}{Theorem}[section]

\newtheorem{lemma}[theorem]{Lemma}

\theoremstyle{definition}

\theoremstyle{remark}

\usepackage[textsize=tiny]{todonotes}

\icmltitlerunning{SafeGround: Know When to Trust GUI Grounding Models via Uncertainty Calibration}

\begin{document}

\twocolumn[

  \icmltitle{\raisebox{-0.3em}{\includegraphics[height=1.5em]{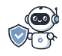}} SafeGround: Know When to Trust GUI Grounding Models \\via Uncertainty Calibration}



  \icmlsetsymbol{equal}{*}

  \begin{icmlauthorlist}
    \icmlauthor{Qingni Wang}{equal,comp,yyy}
    \icmlauthor{Yue Fan}{equal,yyy}
    \icmlauthor{Xin Eric Wang}{comp}
  \end{icmlauthorlist}

  \icmlaffiliation{comp}{University of California, Santa Barbara}
  \icmlaffiliation{yyy}{University of California, Santa Cruz}

  \icmlcorrespondingauthor{Xin Eric Wang}{ericxwang@ucsb.edu}


  \vskip 0.3in
]

\printAffiliationsAndNotice{\icmlEqualContribution}




\begin{abstract}
Graphical User Interface (GUI) grounding aims to translate natural language instructions into executable screen coordinates, enabling automated GUI interaction. Nevertheless, incorrect grounding can result in costly, hard-to-reverse actions (e.g., erroneous payment approvals), raising concerns about model reliability. 
In this paper, we introduce \textsc{SafeGround}, an uncertainty-aware framework for GUI grounding models that enables risk-aware predictions through calibrations before testing.
\textsc{SafeGround} leverages a distribution-aware uncertainty quantification method to capture the spatial dispersion of stochastic samples from outputs of any given model. Then, through the calibration process, \textsc{SafeGround} derives a test-time decision threshold with statistically guaranteed false discovery rate (FDR) control.
We apply \textsc{SafeGround} on multiple GUI grounding models for the challenging ScreenSpot-Pro benchmark.
Experimental results show that our uncertainty measure consistently outperforms existing baselines in distinguishing correct from incorrect predictions, while the calibrated threshold reliably enables rigorous risk control and potentials of substantial system-level accuracy improvements.
Across multiple GUI grounding models, \textsc{SafeGround} improves system-level accuracy by up to 5.38\% percentage points over Gemini-only inference.
\end{abstract}

\begin{figure}[h]
    \centering
    \includegraphics[width=\linewidth]{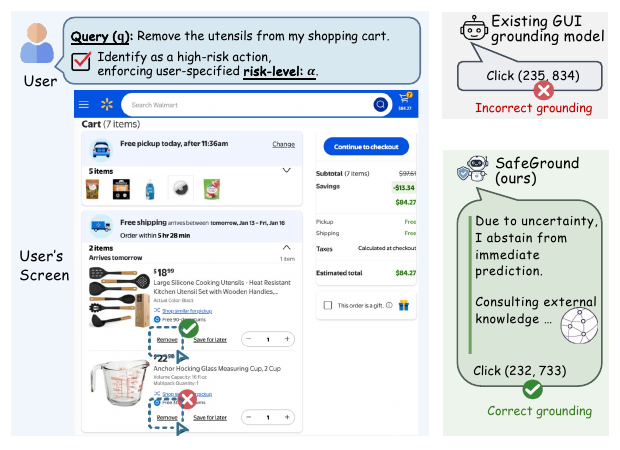}
    \caption{
     While existing models may commit costly errors on hard-to-undo actions (e.g., checkout), \textsc{SafeGround} detects high uncertainty and defers the decision via cascading. This mechanism explicitly limits the risk of erroneous actions to a  user-specified tolerance.%\eric{(1) let's not use shadow in the text. Looks pretty blurring. (2) mark both clicks on the screen.}
    }
    \label{fig:overview}
\end{figure}
\input{section/introduction}
\section{Related Work}
\input{section/relatedwork}

\input{section/method}
\input{section/Experiment}

\input{section/conclusion}
\newpage
\input{section/impact}

\bibliography{example_paper}
\bibliographystyle{icml2026}

\newpage
\appendix
\onecolumn
\input{section/appdendix}

\end{document}

%% file: section/introduction.tex
 \section{Introduction}
 \begin{figure*}[!t]
    \centering
    \includegraphics[width=\textwidth, trim=0cm 5.5cm 0cm 5.5cm,clip]{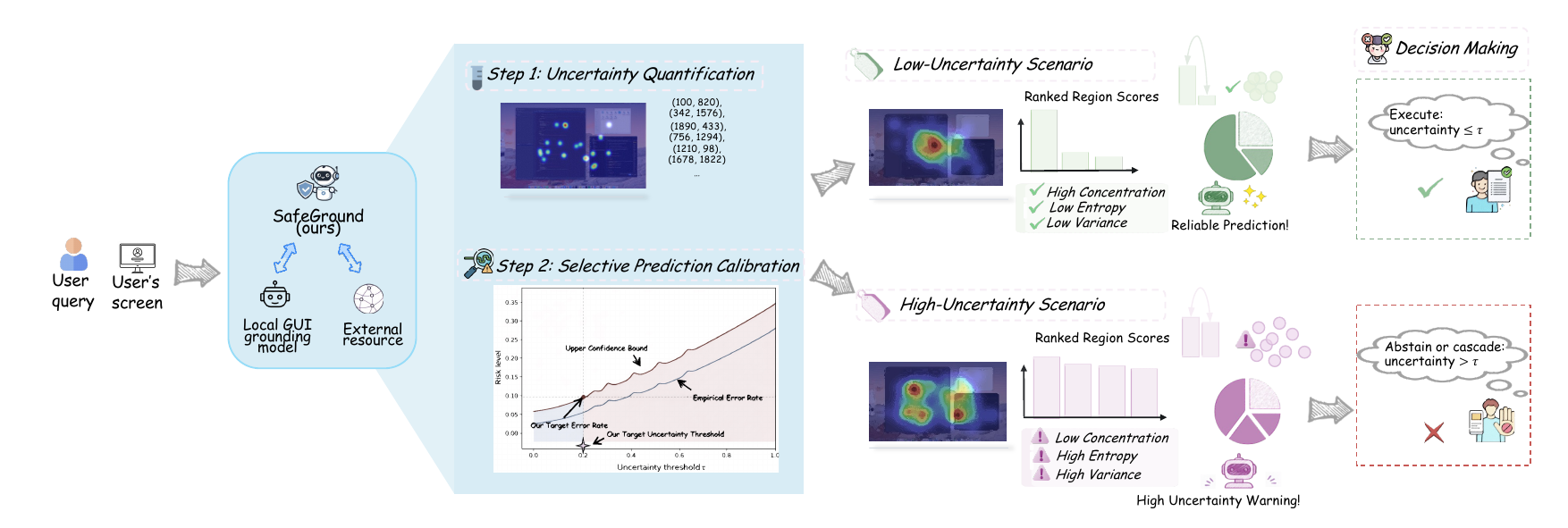}
    \caption{
    Overview of \textsc{SafeGround}.
    Given a GUI input, the model performs multiple stochastic grounding samples to estimate predictive uncertainty.
    An uncertainty threshold $\tau$ is calibrated on a held-out set under a user-specified risk level (i.e, the maximum error rate).
    At test time, predictions with uncertainty $\le \tau$ are executed directly, while high-uncertainty cases are abstained or cascaded.
    Low-uncertainty cases exhibit concentrated region scores, low entropy, and low variance, whereas high-uncertainty cases show dispersed predictions and trigger safety-aware deferral.
    }
    \label{fig:overview2}
\end{figure*}
Graphical User Interface (GUI) grounding is a critical component for autonomous GUI agents, enabling vision-language models (VLMs) to translate natural language instructions into executable screen coordinates~\citep{nguyen2025guiagentssurvey,cheng2024seeclick}. Recent advances have substantially improved grounding accuracy across diverse GUI environments, making it increasingly feasible to deploy such agents in real-world applications~\citep{fan-etal-2025-gui,Hong_2024_CVPR}. 
However, in practical GUI interactions, a single incorrect grounding can trigger costly and hard-to-reverse actions, including erroneous payment approvals or irreversible system configurations~\citep{zhang2025hyperclick}. 
Despite these risks, existing GUI grounding models typically output only point predictions, offering no indication of when a prediction is unreliable or should be deferred~\citep{gawlikowski2022surveyuncertaintydeepneural,hu2023uncertaintynaturallanguageprocessing} as shown in Figure~\ref{fig:overview}.

The aforementioned limitation of existing GUI grounding models motivates the incorporation of uncertainty quantification (UQ) to enable safer decision-making. However, existing UQ techniques are poorly suited for GUI grounding and remain largely underexplored in this setting~\citep{zhang2025hyperclick}. In particular, prior approaches suffer from several key limitations. (1) Uncertainty derived from model probabilities or logits~\citep{hendrycks2017a} assumes access to internal model states, making it infeasible for black-box vision-language models commonly used in GUI agents~\citep{ye2024benchmarking,wang2025sample}.
(2) Verbalized self-assessment~\citep{kadavath2022languagemodelsmostlyknow} relies on strong instruction-following behavior and often fails when models do not explicitly reason about confidence.
(3) Approaches that estimate uncertainty using ground-truth regions, such as~\citet{zhang2025hyperclick}, require annotation and cannot be applied at inference time.
(4) Existing methods focus on producing uncertainty scores alone, without specifying how predictions should be acted upon at deployment time (e.g., whether to accept, defer, or abstain) despite this decision being critical in high-stakes GUI interactions~\citep{geifman2017selective,wang2025coinuncertaintyguardingselectivequestion}.
Collectively, these limitations expose a clear gap between existing UQ approaches and the practical requirements of GUI grounding, where uncertainty must be reliable under limited model access and without test-time supervision~\citep{lin2023generating}.

To address these challenges, we introduce \textsc{SafeGround}, an uncertainty-aware framework that enables risk-aware predictions for existing state-of-the-art GUI grounding models, without requiring access to model internals.
Concretely, as shown in Figure~\ref{fig:overview2}, \textsc{SafeGround} first quantifies the predictive uncertainty of grounding outputs from the spatial distribution of multiple stochastic grounding samples from the same model. 
Then, given the model outputs with estimated uncertainty, we adopt a Learn Then Test (LTT) calibration paradigm to select a decision threshold that rigorously controls the false discovery rate (FDR) of accepted grounding predictions. This calibration procedure provides finite-sample guarantees: with high probability, the proportion of incorrect predictions among all accepted actions does not exceed a user-specified risk level $\alpha$.
At inference time, \textsc{SafeGround} enables a principled selective prediction mechanism. Predictions deemed reliable under the calibrated threshold are executed directly, while high-uncertainty cases are abstained from or deferred to stronger models for further processing. 
Furthermore, with the selective prediction, we realized the cascading inference, where even when the primary model's base accuracy is limited, we can further leverage external resource to aid the prediction, achieving strong system-level accurarcy.

We evaluate \textsc{SafeGround} on the challenging ScreenSpot-Pro benchmark across multiple state-of-the-art GUI grounding models. Experimental results demonstrate that our proposed uncertainty measure consistently outperforms existing baselines in distinguishing correct from incorrect predictions. Especially, \textsc{SafeGround} achieves reliable FDR control in practice and significantly improves overall system accuracy through selective deferral, validating its effectiveness for high-stakes GUI interaction scenarios. Empirically, \textsc{SafeGround} demonstrates clear system-level accuracy gains across different risk levels. For instance, on ScreenSpot-Pro, uncertainty-aware cascading with Holo1.5-7B achieves 58.66\% accuracy at risk level 0.34, improving over Gemini-only inference by 5.38\% points.
% \eric{back with some concrete result comparison.} 
Our contributions can be summarized as follows:
\begin{itemize}
    \item We propose \textsc{SafeGround}, the first framework for uncertainty-aware selective GUI grounding with finite-sample risk guarantees via calibration.
    \item We introduce distribution-aware uncertainty quantification that leverage the spatial dispersion and concentration of stochastic grounding predictions. 
    \item We demonstrate that \textsc{SafeGround} with uncertainty-calibrated selective prediction enables reliable FDR control and improves system-level accuracy in cascading inference on the ScreenSpot-Pro benchmark.
\end{itemize}

%% file: section/relatedwork.tex
\subsection{GUI Grounding}
GUI grounding maps natural language instructions to actionable interface elements or click locations in graphical user interfaces~\citep{nguyen2025guiagentssurvey,fan-etal-2025-gui}. Most existing GUI grounding methods formulate the problem as a text-based coordinate prediction task, where models generate point locations conditioned on the input screenshot and instruction~\citep{chen2023shikraunleashingmultimodalllms,wang2024qwen2vlenhancingvisionlanguagemodels,qin2025uitarspioneeringautomatedgui}.
Recently, motivated by how humans interact with digital interfaces, GUI-Actor introduces an attention-based formulation that aggregates spatial evidence into a single grounding decision~\citep{wu2025gui}. These methods have achieved strong empirical accuracy across diverse GUI environments. However, most existing approaches produce deterministic point predictions and do not explicitly model predictive uncertainty, limiting their ability to assess decision reliability or defer actions under high uncertainty.
\subsection{Uncertainty Estimation}
Uncertainty estimation is widely used to support reliable decision making in AI systems by quantifying the confidence of model predictions~\citep{liu2025uncertaintyquantificationconfidencecalibration}. In large language models, uncertainty has also been derived from probabilistic measures, semantic entropy, or verbalized self-reports~\citep{hou2025probabilisticframeworkllmhallucination, wang2024wordsequenceentropyuncertaintyestimation, 10.1016/j.ijhcs.2025.103455,kuhn2023semanticuncertaintylinguisticinvariances}. In GUI grounding, uncertainty estimation remains largely underexplored. Existing GUI grounding approaches typically rely on probabilistic uncertainty or verbalized uncertainty, both of which have been shown to be systematically miscalibrated, exhibiting a mismatch between predicted confidence and actual grounding accuracy~\citep{zhang2025hyperclick}. This misalignment motivates uncertainty estimation methods that rely solely on model outputs while providing more reliable signals for downstream decision-making, as considered in our work.
\subsection{Learn then Test Calibration}
Learn Then Test (LTT) is a post-hoc calibration paradigm that separates model learning from statistical risk control~\citep{angelopoulos2022learntestcalibratingpredictive}. Given a fixed predictive model, LTT frames decision making as a hypothesis testing problem over a low-dimensional decision space, and uses held-out calibration data to identify parameters that satisfy user-specified risk constraints with finite-sample guarantees. Split conformal prediction (SCP)~\citep{angelopoulos2022gentleintroductionconformalprediction} follows this principle by leveraging data splitting and concentration-based confidence bounds to perform valid risk estimation. Prior work builds on this paradigm to enable reliable decision making in large foundation models~\citep{jung2025trust,wang2025saferriskconstrainedsamplethenfilterlarge,wang2025coinuncertaintyguardingselectivequestion,wang-etal-2025-sconu}. Our approach also builds on the LTT paradigm and extends it to GUI grounding through uncertainty-based calibration of spatial action decisions for the first time.

%% file: section/method.tex
\section{Methodology}
\subsection{Problem Formulation and Notations}
\label{formulation}
Let the GUI grounding model be a function $f: \mathcal{X} \times \mathcal{T} \rightarrow \mathbb{R}^2$, which takes a UI screenshot $x \in \mathcal{X}$ and a user instruction $q \in \mathcal{T}$ as input. 
Given an input pair $(x, q)$, the model predicts a coordinate $\hat{y} = (\hat u,\hat v) \in \mathbb{R}^2$ on the screen. 
Although the model produces a single point prediction, the ground truth for a target UI element is typically provided as a spatial region on the screen, denoted by $B^* \subset  \mathbb{R}^2$. 
A predicted coordinate is considered correct if and only if it falls within the ground-truth region, which we conclude as an admission function $A: \mathbb{R}^2 \times \mathcal{P}(\mathbb{R}^2) \rightarrow \{0,1\}$ with $1$ indicating a correct prediction:
\[
A\big(\hat{y} , B^*\big) = \begin{cases}
1, & \text{if $\hat{y} \in B^*$},\\[4pt]
0, & \text{otherwise.}
\end{cases}
\]
In current coordinate-based GUI grounding models, predictions are deterministic and are not accompanied by explicit uncertainty or confidence estimates, which leaves the trustworthiness of model outputs largely uncharacterized, and may cause users to place unwarranted trust in incorrect predictions, without any indication of potential failure. 
\subsection{Method Overview}
To address this issue, we propose \textsc{SafeGround}, an uncertainty-aware GUI grounding framework that can be integrated with diverse state-of-the-art GUI grounding models without requiring access to internal model states, as illustrated in Figure~\ref{fig:overview2}. \textsc{SafeGround} introduce a user-specified risk level $\alpha \in (0,1)$ that quantifies the maximum tolerable proportion of incorrect predictions, serving as a high-level control signal for how conservatively the system should behave. The risk level $\alpha$ is then translated into an uncertainty threshold $\tau$ through a calibration procedure. Specifically, the GUI grounding model's predictive uncertainty, $U\!\left(\hat{y}^{(\mathrm{MLG})}\right) \in \mathbb{R}$ for a prediction $\hat{y}^{(\mathrm{MLG})}$, is estimated by \textsc{SafeGround} through sampling multiple additional predictions from the GUI grounding model given the same input. The larger values of such uncertainty score indicate lower reliability.
A prediction $\hat{y}$ is correct if $U(\hat{y}) \le \tau$ and rejected otherwise, in which case it is deferred to a stronger model. The threshold $\tau$ is chosen such that, among all admitted predictions, the fraction of incorrect ones, measured by the admission function $A(\hat{y}, B^*)$, is controlled below $\alpha$.

\subsection{Uncertainty Quantification}
\label{3.2}
We first quantify model uncertainty by analyzing the distributional properties of the ranked region scores. Then, three complementary uncertainty measures are introduced, where they are designed to capture complementary failure modes of GUI grounding: local ambiguity among competing targets, global dispersion of belief across regions, and lack of dominant spatial concentration.

\paragraph{Sampling-Based Spatial Distribution Construction}
To move beyond deterministic point predictions and capture the output distribution of GUI grounding models, we employ a Monte Carlo~\citep{pmlr-v48-gal16} sampling strategy followed by spatial aggregation, drawing inspiration from attention-based aggregation mechanisms in~\citep{wu2025gui}. 
Specifically, for each input $(x, q)$, we perform $K$ stochastic forward passes of the grounding model, generating a set of coordinates
$\mathcal{S} = \{\hat{y}^{(i)}\}_{i=1}^K$, where $\hat{y}^{(i)} \in \mathbb{R}^2$.

These sampled coordinates are then projected onto a discretized screen grid to estimate a normalized local density map $P$, which empirically characterizes the spatial distribution of the model’s predictions using only sampled outputs from the model.
Intuitively, high density in a localized area indicates model consistency and thus low uncertainty.
To establish object-level representations, we aggregate connected high-density patches in $P$ into disjoint regions
$\mathcal{R} = \{ R_m \}_{m=1}^M$ through density-based clustering.
Each region $R_m$ is scored by its \textit{average probability density}, denoted as $S_m$, serving as a proxy for the likelihood that the region corresponds to the intended UI element.
Regions are further ranked such that $S_{(1)} \geq S_{(2)} \geq \dots \geq S_{(M)}$.
More implementation details are provided in the Appendix~\ref{Region Construction}.

\paragraph{Uncertainty Measurement 1. Top-Candidate Ambiguity (TA).}
To measure the distinctiveness of a certain prediction from a GUI grounding model, we compute the margin between the two leading candidates. A vanishing margin indicates that the model is uncertain between multiple plausible targets (e.g., two identical exit buttons), therefore, we propose the uncertanty score measured by top-candidate ambiguity:
\begin{equation}
    U_{TA} = \begin{cases} 
    1 - \frac{S_{(1)} - S_{(2)}}{S_{(1)} + \epsilon}, & \text{if } M \ge 2 \\
    \max(0.1,1 - S_{(1)}), & \text{otherwise}
    \end{cases}
\end{equation}
where $\epsilon$ ensures numerical stability. High $U_{TA}$ signifies localized confusion at the decision boundary.

\paragraph{Uncertainty Measurement 2. Informational Dispersion (IE).}
We assess global uncertainty using the entropy of the region score distribution. To ensure a valid probabilistic interpretation, we induce a categorical distribution over the $M$ regions:\begin{equation}\hat{p}_i = \frac{S_{(i)}}{\sum_{j=1}^M S_{(j)}},\end{equation} and then we define the uncertainty score based on information dispersion as the normalized entropy:\begin{equation}U_{\text{IE}} = -\frac{1}{\log M} \sum_{i=1}^M \hat{p}_i \log(\hat{p}_i + \epsilon).\end{equation} Such measurement captures the dispersion of probability mass across regions; a high $U_{\text{IE}}$ indicates that the model's confidence is fragmented, failing to converge on a single consistent hypothesis.

\paragraph{Uncertainty Measurement 3. Concentration Deficit (CD).}
While entropy assesses global disorder, we explicitly quantify the lack of focus with another uncertainty score $U_{CD}$ by examining the quadratic concentration of the distribution:
\begin{equation}
    U_{CD} = 1 - \sum_{i=1}^M \hat{p}_i^2
\end{equation}
Unlike entropy, $U_{CD}$ is more sensitive to the dominance of the top candidates. Higher values of $U_{CD}$ indicate a highly fragmented distribution, suggesting that the model lacks a clear spatial focus and distributes confidence across multiple interface regions.

\paragraph{Combined Uncertainty Score.}
Each uncertainty score captures a distinct aspect of predictive dispersion, and no single measurement is universally dominant across all models and scenarios.
To obtain a unified and deployment-friendly uncertainty signal, we aggregate these three scores into a single one via a fixed weighted combination:
\begin{equation}
    U_{COM}(\hat{y}) = w_{CD} \cdot U_{CD} + w_{IE} \cdot U_{IE} + w_{TA} \cdot U_{TA}.
\end{equation}
We adopt a single set of weights across all models to preserve a plug-and-play interface without model-specific tuning.

\subsection{Uncertainty Calibration for Selective Prediction}
\label{3.3}
Although the proposed uncertainty measures capture predictive uncertainty, they cannot fully distinguish between correct and incorrect predictions. 
To enable user-specified deployment, we further introduce a selective prediction mechanism by calibrating a statistically rigorous decision threshold $\tau$ on the uncertainty score, such that, among all accepted predictions, the proportion of incorrect predictions does not exceed a desired level $\alpha$. 

Following prior SCP-based frameworks, we hold out a calibration set of $N$ data points: $\mathcal{D}_{cal} = \{(x_i, q_i, B_i^*)\}_{i=1}^{N}$. 
For each calibration input pair $(x_i, q_i)$, we produce $\hat{y}_{i}^{(MLG)}$ and quantify its uncertainty score $u_i=U\left(\hat{y}_i^{(MLG)}\right)$. 
Given a candidate threshold $\tau$, we obtain the number of accepted predictions $\sum_i^N \mathbf{1} \{ u_i \leq \tau \}$, and the number of incorrect predictions $\sum_i^N \mathbf{1} \{ u_i \leq \tau, A(\hat{y}_i^{(MLG)} , B_i^*) =0 \}$. 
We then compute the false discovery rate (FDR) on $\mathcal{D}_{cal}$ under threshold $\tau$: 
\begin{equation}
    \mathrm{FDR}_{cal}(\tau) = \frac{\sum_i^N \mathbf{1} \{ u_i \leq \tau, A(\hat{y}_i^{(MLG)} , B_i^*) =0 \}}{\sum_i^N \mathbf{1} \{ u_i \leq \tau \}}
\end{equation}
To provide finite-sample FDR guarantees for the accepted samples at test time, we first introduce an auxiliary lemma. 

\begin{mdframed}[
    hidealllines=true,
    backgroundcolor=lightblue,
    innerleftmargin=3pt,
    innerrightmargin=3pt,
    innertopmargin=4pt,
    innerbottommargin=4pt,
    leftmargin=-3pt,
    rightmargin=-3pt
]
\begin{lemma}[Clopper--Pearson interval~\citep{clopper1934use}]
\label{lem:clopper-pearson}
Let $X \sim \mathrm{Bin}(n,p)$ be the number of successes in $n$ i.i.d.\ Bernoulli trials with success probability $p$.
For any $\delta\in(0,1)$, define the Clopper-Pearson confidence interval
\begin{equation}
\begin{split}
    \Bigl[p_L(X),\, p_U(X)\Bigr]
    \!=\!
    \Bigl[
    & \mathrm{Beta}^{-1}\!\left(\tfrac{\delta}{2};\, X,\, n-X+1\right),\\
    & \mathrm{Beta}^{-1}\!\left(1-\tfrac{\delta}{2};\, X+1,\, n-X\right)
    \Bigr]
\end{split},
\end{equation}
where $\mathrm{Beta}^{-1}(q;a,b)$ denotes the $q$-quantile from a beta distribution with shape parameters $a$ and $b$. 
Then the interval has (at least) nominal coverage:
\begin{equation}
    \mathbb{P}\!\left(p \in [p_L(X),p_U(X)]\right) \ge 1-\delta .
\end{equation}
\end{lemma}\end{mdframed}

In our setting, $X=\sum_i^N \mathbf{1} \{ u_i \leq \tau, A(\hat{y}_i^{(MLG)} , B_i^*) =0 \}$ and $n=\sum_i^N \mathbf{1} \{ u_i \leq \tau \}$. 
Since we focus on controlling the upper tail of the system FDR $R(\tau)$ (thereby constraining test-time FDR), based on Lemma~\ref{lem:clopper-pearson}, we construct a high-probability upper confidence bound, $\hat{\mathrm{FDR}}_{1-\delta}^{upper}(\tau)$, for $R(\tau)$, using its empirical estimate from the calibration data: 
\begin{equation}\label{eq:UCB}
\begin{split}
    \hat{\mathrm{FDR}}_{1-\delta}^{upper}(\tau)&=\mathrm{Beta}\!\left(1-\delta;\, X+1,\, n-X\right)\\
    &=\sup\{R: \Pr (\mathrm{Bin}(n,R)\leq X)\geq \delta\}
\end{split},
\end{equation}
where $\hat{\mathrm{FDR}}_{1-\delta}^{upper}$ guarantees
\begin{equation}\label{eq:UCB guarantee}
    \Pr \big( R(\tau) \leq  \hat{\mathrm{FDR}}_{1-\delta}^{upper}(\tau) \big) \geq 1-\delta.
\end{equation}
Essentially, $\hat{\mathrm{FDR}}_{1-\delta}^{upper}(\tau)$ can be interpreted as the largest plausible value that the system FDR could take, given that an extremely small $\mathrm{FDR}_{cal}(\tau)$ is observed on the calibration set at significance level $\delta$. 
If the true system FDR were to exceed this bound, then observing $\mathrm{FDR}_{cal}(\tau)$ in a single realization would be statistically impossible at the level $\delta$. 
A formal proof of Eq.~\eqref{eq:UCB guarantee} is provided in Appendix~\ref{sec: Proofs}. 

To rigorously constrain test-time FDR, we calibrate $\tau$ such that $\hat{\mathrm{FDR}}_{1-\delta}^{upper}(\tau)$ does not exceed the risk level $\alpha$:
\begin{equation}
    \hat{\tau} = \sup \{ \tau: \hat{\mathrm{FDR}}_{1-\delta}^{upper}(\tau) \leq \alpha \}
\end{equation}
The choice of $\hat{\tau}$ maximizes the acceptance of model predictions (or minimizes the abstention rate), while maintaining marginal FDR control. 
For a test sample $(x_{test}, q_{test}, B_{test}^{*})$ with the model prediction $\hat{y}_{test}^{(MLG)}$ and estimated uncertainty score $u_{test}=U\left(\hat{y}_{test}^{(MLG)}\right)$, by applying the calibrated decision threshold $\hat{\tau}$, we establish the following guarantee
\begin{equation}
    \Pr \left( \Pr \left( A\big(\hat{y}_{test}^{(MLG)} , B^*_{test}\big)=0 \mid u_{test} \leq \hat{\tau} \right) \leq \alpha \right) \geq 1-\delta.
\end{equation}

\paragraph{Cascading Inference.}
At inference time, for each test input $(x_{test}, q_{test})$, we first estimate the model uncertainty $u_{test}$, and then perform selective prediction and escalating:
\begin{itemize}
    \item If $u_{test} \leq \hat{\tau}$, we define the sample as ``safe" and accept the prediction of the primary model.
    \item If $u_{test} > \hat{\tau}$, we flag the sample as ``risky" and escalate the input to a stronger model to enhance performance. 
\end{itemize}

%% file: section/Experiment.tex
\section{Experiment}

\subsection{Experimental Settings}
\paragraph{Models and Dataset}
We conduct our experiments over 6 GUI-grounding models, including Holo1.5~\citep{hai2025holo15modelfamily}, GUI-Actor~\citep{wu2025guiactorcoordinatefreevisualgrounding}, UI-TARS-1.5~\citep{qin2025ui}, GTA1~\citep{yang2025gta1guitesttimescaling}: Holo1.5-3B, Holo1.5-7B, GUI-Actor-2VL-7B, GUI-Actor-2.5VL-7B, UI-TARS-1.5-7B and GTA1-7B.
To assess reliability under high-stakes scenarios, we conduct all experiments on the challenging ScreenSpot-Pro~\citep{li2025screenspotproguigroundingprofessional} benchmark. Additional dataset details are provided in the Appendix~\ref{app:dataset}.
\paragraph{Evaluation Metrics}
To comprehensively evaluate both the discriminative ability of UQ methods and the reliability and effectiveness of \textsc{SafeGround}, we adopt four complementary metrics: Area Under Receiver Operating Characteristic (AUROC), Area Under Accuracy-Rejection Curve (AUARC), FDR, and power~\citep{lin2024generatingconfidenceuncertaintyquantification, wang2025coinuncertaintyguardingselectivequestion}. AUROC measures the ability of uncertainty estimates to distinguish correct from incorrect predictions, while AUARC evaluates whether prediction accuracy improves as high-uncertainty samples are progressively rejected. FDR quantifies the proportion of incorrect predictions among the accepted samples.
Power measures the proportion of correct samples that are retained after uncertainty-based selection, relative to the total number of correct samples. More details about the metrics can be found in Appendix~\ref{metrics}.

\paragraph{Hyperparameters}
For uncertainty estimation, we sample each input 10 times with the decoding temperature set to 1.0 to compute the corresponding UQ score. The most likely generation $\hat{y}_i^{(\mathrm{MLG})}$ is obtained by uniformly sampling one output from the generated candidates.
Specifically, when computing UQ scores, we partition the input into patches with a patch size of 14 to obtain region-level scores $S_i$ for uncertainty estimation.
We repeat the random calibration–test split 100 times and report the mean and standard deviation (mean$\pm$std) over all runs.
All confidence bounds are constructed at a significance level of $\delta = 0.05$.
For the combined uncertainty score $U_{\mathrm{COM}}$, we use a fixed weighting scheme $(w_{CD}, w_{IE}, w_{TA}) = (0.6, 0.2, 0.2)$ across all models.

\subsection{Evaluation of Uncertainty Estimation}

Following prior work~\citep{kuhn2023semantic, band2022benchmarkingbayesiandeeplearning}, we evaluate the quality of uncertainty estimates using AUROC and AUARC, which measure the discriminative ability of uncertainty scores and their effectiveness for selective prediction, respectively. We compare our distribution-aware uncertainty with the probabilistic confidence (PC) baseline, defined as one minus the average token probability~\citep{pouget2016confidence}.

Table~\ref{tab:auroc_comparison} reports AUROC results across six GUI grounding models.
When PC is available, our method consistently achieves higher AUROC.
, and on Holo1.5-7B from 0.6983 to 0.7526.
For models where PC is not directly applicable (e.g., GUI-Actor variants), our method still attains strong AUROC values (up to 0.8155), demonstrating robust error discrimination under limited model access.
Overall, these results suggest that modeling the spatial distribution of grounding predictions yields more informative uncertainty signals than token-level confidence alone.

We further evaluate uncertainty quality using AUARC, which captures accuracy gains as high uncertainty predictions are progressively rejected.
As shown in Table~\ref{tab:auarc_comparison}, our method consistently outperforms baselines across models.
For example, on Holo1.5-3B, AUARC improves from 0.6444 to 0.6576 compared to PC.
These results indicate that our uncertainty estimates are particularly effective for guiding selective prediction decisions.

\begin{figure*}[!t]
    \centering
    \includegraphics[width=0.8\textwidth]{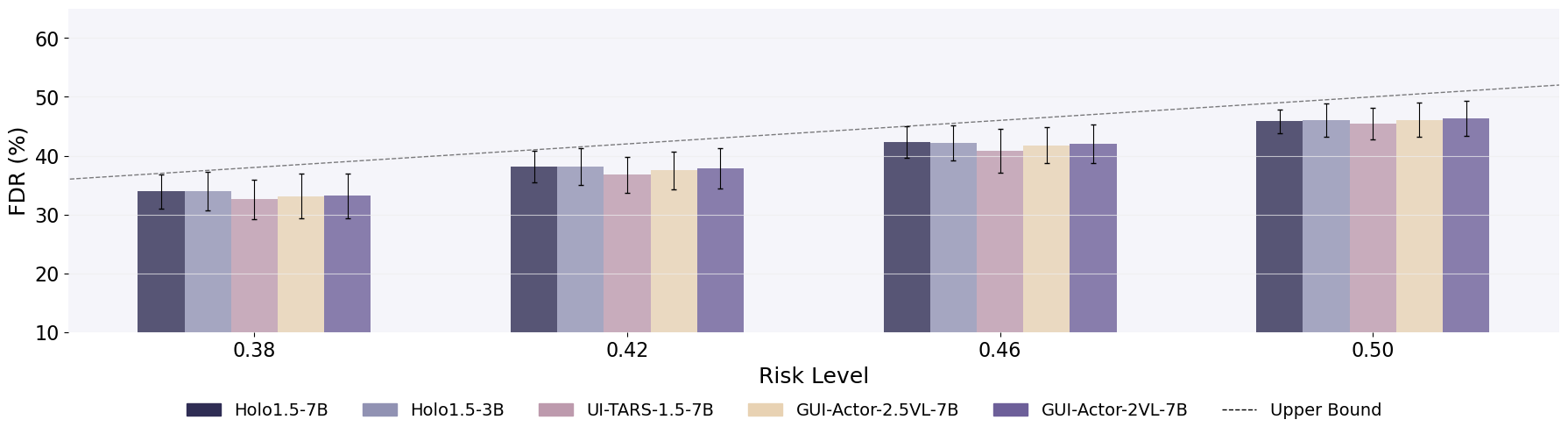}
    \caption{
    Test-time FDR (mean$\pm$std) on the ScreenSpot-Pro dataset under different risk levels.
        }
    \label{fig:guarantee}
\end{figure*}
\begin{figure}[!t]
    \centering
    \includegraphics[width=0.98\columnwidth]{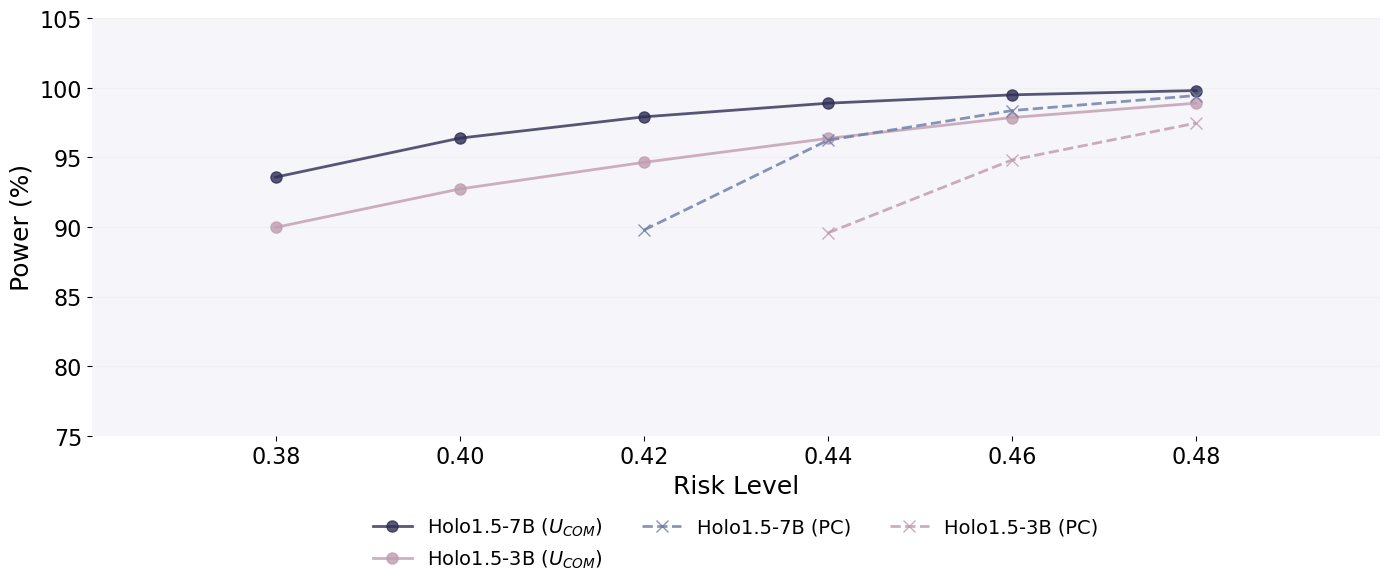}
    \caption{Test-time power (mean) of our $U_{\text{COM}}$ and PC baseline on the ScreenSpot-Pro dataset under different risk levels.}
    \label{fig:power}
\end{figure}
\input{tables/overall}
\input{tables/auroc}
\input{tables/auarc}
\subsection{Selective Prediction with FDR Guarantees}
While AUROC and AUARC evaluate the quality of uncertainty estimates, reliable deployment further requires translating these scores into principled decision rules with explicit risk guarantees. 
We therefore study selective prediction under false discovery rate (FDR) control.

\paragraph{FDR Control Guarantee} 
For each uncertainty method and risk level, we calibrate a decision threshold on the calibration set using the Clopper–Pearson upper confidence bound~\cite{clopper1934use}, ensuring that the test-time FDR does not exceed the specified risk level with high probability.
Figure~\ref{fig:guarantee} illustrates the empirical FDR on the test set across various user-specified risk levels ($\alpha$). Notably, the evaluated risk levels start from a minimum attainable value. This arises because the intrinsic limitations of the base model and the imperfect discriminative power of uncertainty estimates may cause some incorrect predictions to receive relatively low uncertainty scores, making them inseparable from correct ones by thresholding. 
As a result, very stringent FDR requirements may be infeasible to satisfy, as no decision threshold can meet the risk constraint under such conditions~\citep{wang2025saferriskconstrainedsamplethenfilterlarge}.
Importantly, this does not undermine the safety guarantee, as the calibration stage explicitly determines whether a user-specified risk level is achievable prior to deployment, providing a principled fail-safe mechanism for high-stakes interactions.

The results in Figure~\ref{fig:guarantee} show that for all tested models (e.g., Holo1.5, UI-TARS), the actual FDR is consistently bounded below the theoretical upper bound. This empirically verifies that \textsc{SafeGround} provides rigorous safety guarantees, ensuring that, with high probability, the error rate among accepted predictions is controlled at the specified level.

\paragraph{Power Comparison}
In addition to FDR, we report power to further characterize the effectiveness of selective prediction. Higher power indicates that the uncertainty estimates more precisely identify truly risky cases, allowing the system to retain a larger set of reliable predictions without violating the target FDR. Figure~\ref{fig:power} compares the power of our method $U_{\text{COM}}$ versus the PC baseline under identical risk levels. 
Across the evaluated models, $U_{\text{COM}}$ demonstrates superior robustness, particularly at strict risk levels (e.g., 0.38) where PC often fails to yield valid predictions. 
Notably, the minimum attainable risk level at which PC can satisfy the FDR constraint is consistently higher than that of $U_{\text{COM}}$, indicating a narrower feasible operating range for PC.
$U_{\text{COM}}$ consistently outperforms PC, retaining a significantly larger volume of correct responses.
These results indicate that $U_{\text{COM}}$ is systematically less conservative than PC: it accepts a larger fraction of correct predictions while still satisfying the same FDR constraint.

\subsection{Cascading Inference}
Finally, we study the system-level benefits of uncertainty-aware decision making in a cascaded inference setting. 
Given that powerful external models (e.g., Gemini) often incur latency and financial costs, our goal is to improve system accuracy by selectively invoking stronger models when the uncertainty of the base model exceeds a calibrated threshold.
Specifically, we fix the calibration split ratio to 0.2 and use the remaining 80\% of the data as the test set to evaluate the cascaded system.
At test time, predictions with uncertainty scores below or equal to the threshold are handled by the primary local grounding model, while high-uncertainty cases are deferred to the stronger expert model, Gemini-3-pro~\citep{team2023gemini}.

\begin{figure}[!t]
    \centering
    \includegraphics[width=0.98\columnwidth]{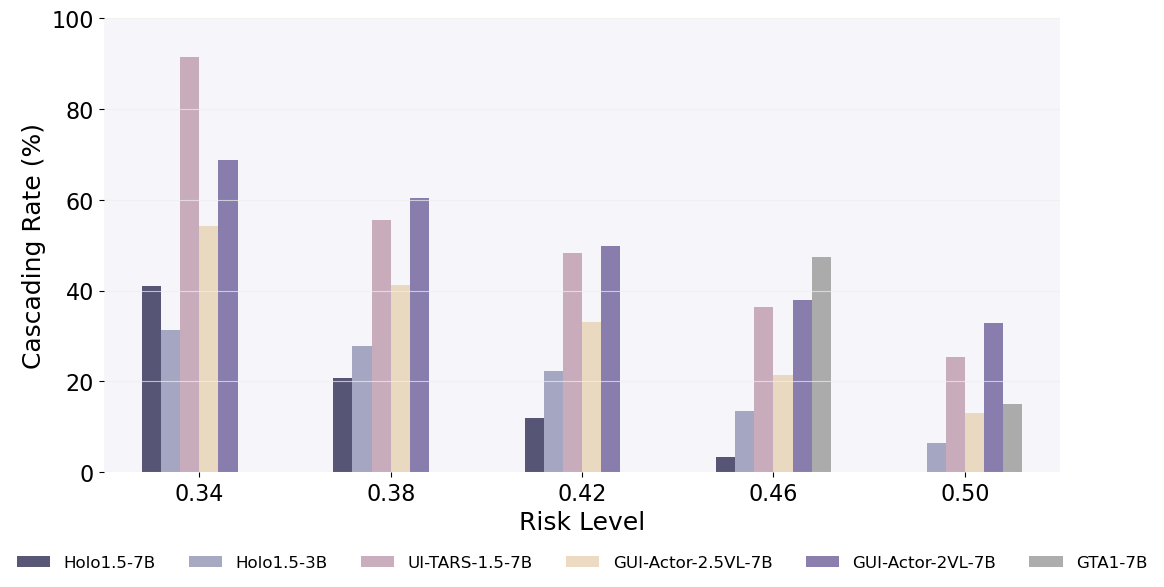}
    \caption{Cascading rate (fraction of test samples deferred to Gemini) across different risk levels.}
    \label{fig:cascading}
\end{figure}

Table~\ref{tab:accuracy_with_uncertainty_gemini} reports the accuracy of uncertainty-aware Gemini cascading under different risk levels. Across a wide range of feasible risk levels, the proposed approach consistently improves system accuracy over both Gemini-only inference and the base models, demonstrating the effectiveness of uncertainty-aware cascading. At relatively small risk levels, uncertainty-aware cascading yields substantial accuracy gains. For instance, with Holo1.5-7B at risk level $0.34$, the system achieves $58.66\%$ accuracy, outperforming Gemini-only inference by $5.38\%$.
As the risk level increases, the improvement gradually diminishes, since fewer high-uncertainty samples are deferred to Gemini, and the system behavior approaches that of the base model.
The effect is more pronounced for models such as Holo1.5-3B and UI-TARS-1.5-7B, where uncertainty-aware cascading improves accuracy by more than $7\%$ to $13\%$ over the base models at relatively small risk levels.
We also report the cascading rate in Figure~\ref{fig:cascading}, i.e., the fraction of test samples deferred to Gemini.
As the risk level increases, the cascading rate consistently decreases across all models, indicating that fewer uncertain cases are escalated to the expert model.
This reflects the inherent trade-off between accuracy and expert invocation cost in uncertainty-aware cascading.

\begin{figure}[!t]
\centering
\begin{minipage}[t]{0.48\columnwidth}
    \centering
    \includegraphics[width=\linewidth]{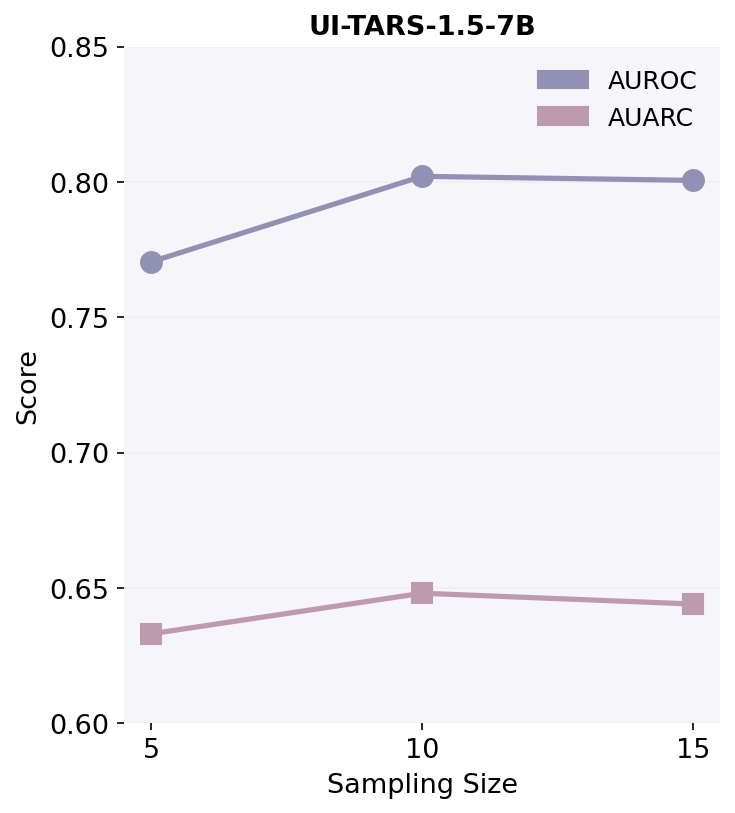}
    \captionof{figure}{Effect of sampling size $K$ on uncertainty estimation quality for UI-TARS-1.5-7B.}
    \label{fig:sampling_efficiency}
\end{minipage}
\hfill
\begin{minipage}[t]{0.48\columnwidth}
    \centering
    \includegraphics[width=\linewidth]{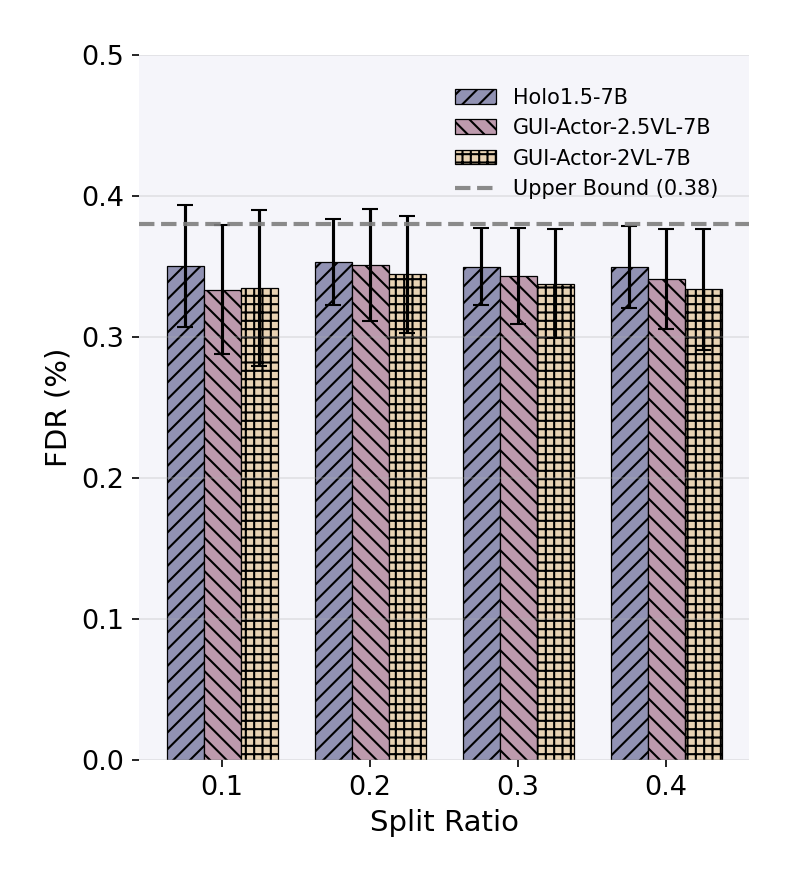}
    \captionof{figure}{Test-time FDR results of various calibration test split ratios.}
    \label{fig:splitratio}
\end{minipage}
\end{figure}

\subsection{Sensitivity Analyses}

\paragraph{Sampling Efficiency}
We investigate the trade-off between computational cost and estimation quality by varying the sample count $K$ and measuring the resulting AUROC and AUARC. As shown in Figure~\ref{fig:sampling_efficiency}, increasing the sample size from $K=5$ to $K=10$ yields a improvement for both metrics, indicating that the proposed uncertainty estimates are already effective with a small number of samples. In contrast, further increasing $K$ from 10 to 15 leads to only marginal changes. Based on this trade-off between performance and computational cost, we set $K=10$ as the default sampling size in all experiments.

\paragraph{Ablation of Uncertainty Components}
We analyze the contribution of individual uncertainty components, $U_{TA}$ , $U_{IE}$, and $U_{CD}$, across different GUI grounding models.
As shown in Table~\ref{tab:ablation_components}, the most informative uncertainty cue is model-dependent.
On GTA1, $U_{TA}$ is the strongest single signal, whereas for GUI-Actor-2VL and Holo1.5, $U_{CD}$ is more effective, and $U_{TA}$ alone is insufficient.

Across all models, no single component consistently dominates.
$U_{COM}$ achieves stable performance in all settings, and removing the dominant component for a given model leads to a clear drop in both AUROC and AUARC.
This indicates that combining complementary cues yields a more robust, model-agnostic uncertainty estimate for selective prediction.
Additional robustness analyses with respect to the uncertainty weighting are provided in the Appendix~\ref{weightsensitivity}.
\input{tables/components}

\paragraph{Sensitivity to Calibration-Test Split Ratio}
We further study the sensitivity of our method to the calibration–test split ratio when using the combined uncertainty measure \(U_{\text{COM}}\).
Specifically, we vary the proportion of data allocated to the calibration set while keeping the target risk level fixed, and evaluate the resulting empirical FDR on the test set.
As shown in Figure~\ref{fig:splitratio}, across a wide range of split ratios, the empirical FDR achieved by all three models remains consistently below the target upper bound.
These results suggest that our approach does not rely on a carefully tuned split ratio and can be applied robustly in practical settings.

%% file: tables/overall.tex
\begin{table}[t]
\centering
\setlength\tabcolsep{2pt}

\caption{System-level accuracy ($\%$) of uncertainty-calibrated cascading under different risk levels.
``--'' indicates infeasible risk levels.
Parentheses show $\Delta$ over the corresponding model baseline (no cascading).
All reported accuracies are computed on the test split, with a test ratio of 0.8.}
\label{tab:accuracy_with_uncertainty_gemini}

\resizebox{0.5\textwidth}{!}{
\begin{tabular}{l ccccc}
\toprule
 & \multicolumn{5}{c}{\textbf{Risk Level}} \\
\cmidrule(lr){2-6}
\textbf{Model}
& \textbf{0.34} & \textbf{0.38} & \textbf{0.42} & \textbf{0.46} & \textbf{0.50} \\
\midrule

Gemini-only & \multicolumn{5}{>{\columncolor{gray!10}}l}{ 53.28} \\
\addlinespace \addlinespace

Holo1.5-7B
& \multicolumn{5}{>{\columncolor{gray!10}}l}{ 52.41} \\
{\footnotesize\textbf{\textsc{(+SafeGround)}}}
& \textbf{58.66} \gainG{+ 6.25}
& 57.87 \gainG{+ 5.46}
& 55.73 \gainG{+ 3.32}
& 53.20 \gainG{+ 0.79}
& 52.41 \gainG{+ 0.00} \\
\addlinespace \addlinespace

Holo1.5-3B
& \multicolumn{5}{>{\columncolor{gray!10}}l}{ 45.45} \\
{\footnotesize\textbf{\textsc{(+SafeGround)}}}
& \textbf{53.44} \gainG{+ 7.99}
& 52.73 \gainG{+ 7.28}
& 52.02 \gainG{+ 6.57}
& 49.25 \gainG{+ 3.80}
& 47.35 \gainG{+ 1.90} \\
\addlinespace \addlinespace

UI-TARS-1.5-7B
& \multicolumn{5}{>{\columncolor{gray!10}}l}{ 41.58} \\
{\footnotesize\textbf{\textsc{(+SafeGround)}}}
& 53.68 \gainG{+12.10}
& \textbf{54.70} \gainG{+13.12}
& 53.04 \gainG{+11.46}
& 50.43 \gainG{+ 8.85}
& 47.91 \gainG{+ 6.33} \\
\addlinespace \addlinespace

GUI-Actor-2.5VL-7B
& \multicolumn{5}{>{\columncolor{gray!10}}l}{ 45.69} \\
{\footnotesize\textbf{\textsc{(+SafeGround)}}}
& \textbf{55.18} \gainG{+ 9.49}
& 54.86 \gainG{+ 9.17}
& 53.60 \gainG{+ 7.91}
& 51.38 \gainG{+ 5.69}
& 49.17 \gainG{+ 3.48} \\
\addlinespace \addlinespace

GUI-Actor-2VL-7B
& \multicolumn{5}{>{\columncolor{gray!10}}l}{ 40.79} \\
{\footnotesize\textbf{\textsc{(+SafeGround)}}}
& \textbf{55.18} \gainG{+14.39}
& 53.28 \gainG{+12.49}
& 53.99 \gainG{+13.20}
& 52.96 \gainG{+12.17}
& 50.67 \gainG{+9.88} \\
\addlinespace \addlinespace

GTA1-7B
& \multicolumn{5}{>{\columncolor{gray!10}}l}{ 46.88} \\
{\footnotesize\textbf{\textsc{(+SafeGround)}}}
& --
& --
& --
& \textbf{53.12} \gainG{+ 6.24}
& 49.96 \gainG{+ 3.08} \\

\bottomrule
\end{tabular}
}
\end{table}

%% file: tables/auroc.tex
\begin{table}[t]
     \centering
     \setlength\tabcolsep{8pt}

    \caption{AUROC comparison of uncertainty quantification methods across different models. 
    The best results for each model are highlighted in \textbf{bold}. PC is the Probabilistic Confidence baseline.}
    \label{tab:auroc_comparison}
    \adjustbox{max width=0.7\linewidth}{    
    \begin{tabular}{l|cc}
        \toprule
        \textbf{Model} 
        & \multicolumn{2}{c}{\textbf{Uncertainty Score}} \\
        \cmidrule(lr){2-3}
        & \textit{PC} 
        & \textbf{$U_{\text{COM}}$ (Ours)} \\
        \midrule
        Holo1.5-3B         
            & 0.7576 
            & \cellcolor{LightPurple}\textbf{0.8056} \\
        Holo1.5-7B         
            & 0.6983 
            & \cellcolor{LightPurple}\textbf{0.7526}  \\
        GUI-Actor-2.5VL-7B 
            & -     
            & \cellcolor{LightPurple}\textbf{0.7793} \\
        UI-TARS-1.5-7B     
            & 0.7844 
            & \cellcolor{LightPurple}\textbf{0.8021}  \\
        GUI-Actor-2VL-7B   
            & -     
            & \cellcolor{LightPurple}\textbf{0.8155} \\
        GTA1-7B            
            & 0.6114 
            & \cellcolor{LightPurple}\textbf{0.6344} \\
        \bottomrule
    \end{tabular}
    }
\end{table}

%% file: tables/auarc.tex
\begin{table}[h]
    \centering
    \footnotesize
    \caption{AUARC comparison of uncertainty quantification methods across different models. 
    The best results for each model are highlighted in \textbf{bold}.}
    \label{tab:auarc_comparison}
    \adjustbox{max width=0.9\linewidth}{    
    \begin{tabular}{l|ccc}
        \toprule
        \textbf{Model} 
        & \multicolumn{3}{c}{\textbf{Uncertainty Score}} \\
        \cmidrule(lr){2-4}
        & \textit{Random} 
        & \textit{PC} 
        & \textbf{$U_{\text{COM}}$ (Ours)} \\
        \midrule
        Holo1.5-3B         
            & 0.4706 
            & 0.6444 
            & \cellcolor{LightPurple}\textbf{0.6576}  \\
        Holo1.5-7B         
            & 0.5345 
            & 0.6686 
            & \cellcolor{LightPurple}\textbf{0.6705}  \\
        GUI-Actor-2.5VL-7B 
            & 0.4662 
            & --     
            & \cellcolor{LightPurple}\textbf{0.7156} \\
        GUI-Actor-2VL-7B   
            & 0.4130 
            & --     
            & \cellcolor{LightPurple}\textbf{0.7166}  \\
        UI-TARS-1.5-7B     
            & 0.4231 
            & 0.6222 
            & \cellcolor{LightPurple}\textbf{0.6480}  \\
        GTA1-7B            
            & 0.4769 
            & \textbf{0.5521} 
            & 0.5511 \\
        \bottomrule
    \end{tabular}
    }
\end{table}

%% file: tables/components.tex
\begin{table}[t]
    \centering
    \caption{
    Ablation study of uncertainty components on GTA1, GUI-Actor-2VL, and Holo1.5 models.
    Best results within each model block are highlighted in bold.
    }
    \label{tab:ablation_components}
    \adjustbox{max width=0.75\linewidth}{    
    \begin{tabular}{@{}llcc@{}}
        \toprule
        \textbf{Model} & \textbf{Uncertainty} & \textbf{AUROC} & \textbf{AUARC} \\
        \midrule

        \multirow{5}{*}{GTA1}
        & $U_{TA}$              & 0.6228 & 0.5481 \\
        & $U_{IE}$             & 0.5916 & 0.5390 \\
        & $U_{CD}$       & 0.5917 & 0.5389 \\
        & \textbf{$U_{COM}$}   & \textbf{0.6344} & \textbf{0.5511} \\
        & w/o $U_{TA}$          & 0.5917 & 0.5389 \\
        \cmidrule(lr){1-4}

        \multirow{5}{*}{GUI-Actor-2VL-7B}
        & $U_{TA}$              & 0.4844 & 0.4335 \\
        & $U_{IE}$             & 0.7731 & 0.6435 \\
        & $U_{CD}$       & 0.7894 & 0.6505 \\
        & \textbf{$U_{COM}$}   & \textbf{0.8155} & \textbf{0.7166} \\
        & w/o $U_{CD}$   & 0.7987 & 0.6221 \\
        \cmidrule(lr){1-4}

        \multirow{5}{*}{Holo1.5-7B}
        & $U_{TA}$              & 0.6296 & 0.6284 \\
        & $U_{IE}$             & 0.7380 & 0.6670 \\
        & \textbf{$U_{CD}$} & \textbf{0.7529} & \textbf{0.6716} \\
        & $U_{COM}$             & 0.7526 & 0.6705 \\
        & w/o $U_{CD}$    & 0.7303 & 0.6483 \\

        \bottomrule
    \end{tabular}}
\end{table}

%% file: section/conclusion.tex
\section{Conclusion}
We presented \textsc{SafeGround}, an uncertainty-aware framework that enables reliable and risk controlled GUI grounding under limited model access.
By modeling spatial uncertainty from stochastic grounding samples, \textsc{SafeGround} captures distributional signals that go beyond point predictions and provide effective discrimination between correct and incorrect predictions.
Based on uncertainty estimation, we further calibrate decision thresholds with finite-sample guarantees, supporting deployment-time decision making in high-stakes GUI interactions.
Extensive experiments demonstrate that \textsc{SafeGround} achieves accurate uncertainty discrimination, rigorous FDR control, and improved system-level performance through selective prediction and cascading inference. We hope this work provides a principled way for deploying GUI agents with safety guarantees.
% principled foundation

%% file: section/impact.tex
\section*{Impact Statement}
This paper introduces \textsc{SafeGround}, a framework that significantly enhances the reliability and safety of autonomous GUI agents. By providing the first principled method for uncertainty quantification in GUI grounding with finite-sample statistical guarantees, our work addresses a critical bottleneck in the real-world deployment of visual agents, the risk of high-stakes, irreversible errors (e.g., erroneous financial transactions). Beyond improving individual model reliability, the proposed selective deferral mechanism demonstrates that local models, when combined with uncertainty-aware cascading to powerful external experts, can achieve superior system-level accuracy with substantially reduced computational costs. 
This research provides a foundational step toward trustworthy human-AI interaction in digital environments, ensuring that automated systems ``know when they don't know" and make conservative decisions under ambiguous conditions.

%% file: section/appdendix.tex
\section*{Limitation}
Our uncertainty estimation relies on the variability in the sampled predictions to characterize spatial ambiguity.
For highly deterministic models with limited sampling diversity, the resulting spatial distributions may be less informative.
Despite these limitations, \textsc{SafeGround} provides a general and principled foundation for uncertainty-aware GUI grounding.

\input{section/Appendix/a}

\input{section/Appendix/b}
\input{section/Appendix/c}
\input{section/Appendix/d}
\input{section/Appendix/e}
\input{section/Appendix/f}

%% file: section/Appendix/a.tex
\section{Proofs}
\label{sec: Proofs}
In this section, we provide a compelete proof that the upper confidence bound $\hat{\mathrm{FDR}}_{1-\delta}^{upper}(\tau)$ defined in Eq.~\eqref{eq:UCB} satisfies the statistical guarantee in Eq.~\eqref{eq:UCB}. 
Recall $\hat{\mathrm{FDR}}_{1-\delta}^{upper}(\tau)=\sup\{R: \Pr (\mathrm{Bin}(n,R)\leq X)\geq \delta\}$, where $n=\sum_i^N \mathbf{1} \{ u_i \leq \tau \}$ is the number of accepted calibration samples, and $X=\sum_i^N \mathbf{1} \{ u_i \leq \tau, A(\hat{y}_i^{(MLG)} , B_i^*) =0 \}$ is the number of accepted incorrect calibration samples. 
In general, $\mathrm{Bin}(n,R)$ denotes the random variable representing the number of successes in $n$ Bernoulli trials when the system success probability is $R$. 
In our setting, it corresponds to the random variable counting the number of errors among $n$ samples when the system FDR is $R$ under a given threshold $\tau$. 

We define the cumulative distribution function (CDF) of the random variable $\hat{R}(\tau)=\frac{\mathrm{Bin}(n;R(\tau))}{n}$, corresponding to the error rate over any $n$ accepted samples when the system FDR is $R(\tau)$, as
\begin{equation}
    \mathrm{CDF} \big( r \mid R(\tau) \big) = \Pr \big( \hat{R}(\tau) \leq r \mid R(\tau)  \big).
\end{equation}
By the definition of $\hat{\mathrm{FDR}}_{1-\delta}^{upper}(\tau)$, we have 
\begin{equation}
    \mathrm{CDF}\left( \frac{X}{n} \mid \hat{\mathrm{FDR}}_{1-\delta}^{upper}(\tau) \right)=\delta.
\end{equation}
If $R(\tau) > \hat{\mathrm{FDR}}_{1-\delta}^{upper}(\tau)$, we have $\mathrm{CDF}\left( \frac{X}{n} \mid R(\tau) \right) \leq \delta$. 
Then, we have 
\begin{equation}
\begin{split}
    \Pr \left( R(\tau) \leq \hat{\mathrm{FDR}}_{1-\delta}^{upper}(\tau) \right) &= 1 - \Pr \left( R(\tau) > \hat{\mathrm{FDR}}_{1-\delta}^{upper}(\tau) \right)\\
    & \geq 1-\Pr \left( \mathrm{CDF}\left( \frac{X}{n} \mid R(\tau) \right) \leq \delta \right)
\end{split}. 
\end{equation}
We further the Inverse Cumulative Distribution Function (ICDF): 
\begin{equation}
    \mathrm{CDF}^{-1} \big(p \mid R(\tau) \big) = \sup \left\{ r :  \mathrm{CDF}\left( r \mid R(\tau) \right) \leq p \right\}. 
\end{equation}
If $\mathrm{CDF}\left( \frac{X}{n} \mid R(\tau) \right) \leq \delta$, we have $\frac{X}{n} \leq \mathrm{CDF}^{-1} \big(\delta \mid R(\tau) \big) $. 
We then obtain 
\begin{equation}
    \Pr \left( R(\tau) \leq \hat{\mathrm{FDR}}_{1-\delta}^{upper}(\tau) \right) \geq 1 - \Pr \left( \frac{X}{n} \leq \mathrm{CDF}^{-1} \big(\delta \mid R(\tau) \big) \right) .
\end{equation}
Since $\frac{X}{n}$ is exactly the empirical error rate observed over the $n$ accepted samples in the calibration set, the probability that it is less than or equal to $\mathrm{CDF}^{-1} \big(\delta \mid R(\tau) \big)$ does not exceed $\delta$. 
Finally, we conclude 
\begin{equation}
    \Pr \left( R(\tau) \leq \hat{\mathrm{FDR}}_{1-\delta}^{upper}(\tau) \right) \geq 1 -  \delta. 
\end{equation}
In this way, we obtain an upper bound on the system FDR at threshold $\tau$ with at least $1-\delta$ confidence. 
At test time, by the exchangeability condition, we provide marginal guarantees of FDR control.

%% file: section/Appendix/b.tex
\section{Details of Experimental Settings}
\subsection{Dataset}
\label{app:dataset}
\paragraph{ScreenSpot-Pro}
ScreenSpot-Pro consists of 1581 UI screenshots paired with natural language instructions that refer to target UI elements on the screen.
Each target is annotated as a spatial region rather than a single point.
Compared to earlier GUI grounding benchmarks, ScreenSpot-Pro features higher visual complexity, denser UI layouts, and more fine-grained distinctions between neighboring elements, making it particularly suitable for studying uncertainty-aware grounding.
\subsection{Evaluation Metrics}
\label{metrics}
We evaluate uncertainty estimation quality and selective prediction performance using four complementary metrics: AUROC, AUARC, FDR, and power.
All metrics are defined with respect to the admission function 
$A(\hat{y}, B^*) \in \{0,1\}$ introduced in Section~\ref{formulation}, which indicates whether a grounding prediction is admissible.

\paragraph{Area Under Receiver Operating Characteristic (AUROC)}
Let $U(\hat{y})$ denote an uncertainty score, where larger values indicate higher uncertainty.
AUROC measures how well $U(\hat{y})$ separates inadmissible predictions from admissible ones.
Formally, AUROC is the area under the receiver operating characteristic curve obtained by thresholding $U(\hat{y})$ to predict whether $A(\hat{y}, B^*) = 0$.
A higher AUROC indicates stronger discriminative ability of the uncertainty estimate.

\paragraph{Area Under Accuracy-Rejection Curve (AUARC)}
AUARC evaluates selective prediction behavior by measuring how accuracy changes as predictions with high uncertainty are rejected.
Let $\mathcal{S}_\tau = \{ i : U(\hat{y}_i) \le \tau \}$ denote the set of accepted samples under threshold $\tau$.
The accuracy at $\tau$ is defined as
\[
\mathrm{Acc}(\tau) = \frac{1}{|\mathcal{S}_\tau|} \sum_{i \in \mathcal{S}_\tau} A(\hat{y}_i, B_i^*).
\]
In practice, $\tau$ is chosen to correspond to a target rejection rate, and AUARC is computed as the area under the curve of $\mathrm{Acc}(\tau)$ as a function of the rejection rate.

\paragraph{False Discovery Rate (FDR)}
Under a given uncertainty threshold $\tau$, the false discovery rate is defined as
\[
\mathrm{FDR}(\tau) =
\frac{\sum_i \mathbb{I}\big(U(\hat{y}_i) \le \tau \big)\, \mathbb{I}\big(A(\hat{y}_i, B_i^*) = 0\big)}
{\sum_i \mathbb{I}\big(U(\hat{y}_i) \le \tau \big)}.
\]
FDR quantifies the proportion of inadmissible predictions among all accepted predictions and serves as the primary risk metric controlled by \textsc{SafeGround}.

\paragraph{Power}
Power measures the proportion of correct predictions retained by selective prediction under a risk constraint and is defined as:
\[
\mathrm{Power}(\tau) = \frac{\sum_{i=1}^N \mathbb{I}\big(U(\hat{y}_i) \le \tau \big) \mathbb{I}\big(A(\hat{y}_i, B_i^*) = 1\big)}{\sum_{i=1}^N \mathbb{I}\big(A(\hat{y}_i, B_i^*) = 1\big)} .
\]
Higher power indicates that more correct predictions are retained while satisfying the specified FDR constraint.

\subsection{Spatial Region Construction}
\label{Region Construction}
Given an input image--instruction pair $(x,q)$, we obtain a set of $K$ sampled grounding predictions
$\mathcal{S}=\{\hat{y}^{(i)}=(x^{(i)},y^{(i)})\}_{i=1}^{K}$ via stochastic decoding.
To lift these point-wise samples into a spatial distribution, we discretize the screen into a fixed
$H\times W$ grid of patches and map each sampled coordinate to its corresponding patch.
Let $C_{u,v}$ denote the number of samples falling into patch $(u,v)$.
We then normalize the resulting count map to obtain a spatial probability distribution
\begin{equation}
P_{u,v} = \frac{C_{u,v}}{\sum_{u',v'} C_{u',v'}} ,
\end{equation}
which serves as an empirical estimate of the model’s predictive density over the output space.

\paragraph{Region Extraction}
To identify object-level grounding hypotheses, we first filter low-density patches using an
instance-adaptive threshold.
Specifically, let $P_{\max}=\max_{u,v} P_{u,v}$, and retain only patches satisfying
$P_{u,v} > \beta P_{\max}$, where $\beta$ is a fixed ratio (set to $0.3$ in our experiments, following~\citep{wu2025gui}).
We then group spatially adjacent retained patches (using 4-connected neighborhood) into connected components.
This yields a set of disjoint regions
$\mathcal{R}=\{R_m\}_{m=1}^{M}$, each corresponding to a plausible grounding target.

\paragraph{Region Scoring}
For each region $R_m$, we compute a region-level score
\begin{equation}
S_m = \frac{1}{|R_m|} \sum_{(u,v)\in R_m} P_{u,v},
\end{equation}
i.e., the average probability density within the region.
This score reflects the relative support assigned to the region by the sampled predictions while
remaining invariant to region size.
The resulting region scores $\{S_m\}_{m=1}^{M}$ are subsequently normalized and used to compute
the uncertainty metrics described in Section~\ref{3.2}.

%% file: section/Appendix/c.tex
\section{Threshold Calibration with Finite-Sample Guarantees}

This section details the threshold calibration procedure used in SafeGround to obtain finite-sample guarantees on selective prediction risk, based on Clopper–Pearson confidence bounds, as summarized in Algorithm~\ref{alg:safeground_cp_calibration}.

\begin{algorithm}[t]
\caption{SafeGround: Clopper--Pearson Threshold Calibration with Sampling-Based Spatial Uncertainty}
\label{alg:safeground_cp_calibration}
\begin{algorithmic}[1]
\STATE \textbf{Input:} GUI grounding model $f$; calibration set $\mathcal{D}_{cal}=\{(x_i,q_i,B_i^*)\}_{i=1}^{N}$;
sample count $K$; patch grid size $H\times W$; region threshold ratio $\beta$;
admission function $A(\hat{y},B^*)$; risk level $\alpha$; significance level $\delta$;
weights $(w_{CD},w_{IE},w_{TA})$
\STATE \textbf{Output:} calibrated uncertainty threshold $\hat{\tau}$

\FOR{$i=1$ \textbf{to} $N$}
    \STATE \textbf{(Primary prediction)} Obtain $\hat{y}^{(MLG)}_i \leftarrow f(x_i,q_i)$
    \STATE \textbf{(Sampling)} Draw $K$ stochastic predictions $\mathcal{S}_i=\{\hat{y}^{(k)}_i\}_{k=1}^{K}$ via stochastic decoding
    \STATE \textbf{(Discretized density map)} Initialize count map $C\in\mathbb{N}^{H\times W}\leftarrow 0$
    \FOR{$k=1$ \textbf{to} $K$}
        \STATE Map $\hat{y}^{(k)}_i$ to patch index $(u,v)$ and set $C_{u,v}\leftarrow C_{u,v}+1$
    \ENDFOR
    \STATE Normalize to density $P_{u,v}\leftarrow \frac{C_{u,v}}{\sum_{u',v'} C_{u',v'}}$

    \STATE \textbf{(Region extraction)} $P_{\max}\leftarrow \max_{u,v} P_{u,v}$; mask $M_{u,v}\leftarrow \mathbb{I}\{P_{u,v}>\beta P_{\max}\}$
    \STATE Group 4-connected active patches in $M$ into $M_i$ connected components via BFS,
           yielding regions $\mathcal{R}_i=\{R_{i,m}\}_{m=1}^{M_i}$

    \STATE \textbf{(Region scoring)} For each region $R_{i,m}$, compute
           $S_{i,m}\leftarrow \frac{1}{|R_{i,m}|}\sum_{(u,v)\in R_{i,m}} P_{u,v}$
    \STATE Sort scores in descending order: $S_{i,(1)}\ge \cdots \ge S_{i,(M_i)}$
    \STATE Induce categorical distribution $\hat{p}_{i,j}\leftarrow \frac{S_{i,(j)}}{\sum_{\ell=1}^{M_i} S_{i,(\ell)}}$

    \STATE \textbf{(Uncertainty components)}
    \STATE $U_{TA,i}\leftarrow
    \begin{cases}
    1-\frac{S_{i,(1)}-S_{i,(2)}}{S_{i,(1)}+\epsilon}, & M_i\ge 2\\
    \max(0.1,\,1-S_{i,(1)}), & M_i=1
    \end{cases}$
    \STATE $U_{IE,i}\leftarrow -\frac{1}{\log M_i}\sum_{j=1}^{M_i}\hat{p}_{i,j}\log(\hat{p}_{i,j}+\epsilon)$
    \STATE $U_{CD,i}\leftarrow 1-\sum_{j=1}^{M_i}\hat{p}_{i,j}^{2}$
    \STATE \textbf{(Combined uncertainty)} $u_i \leftarrow w_{CD}U_{CD,i}+w_{IE}U_{IE,i}+w_{TA}U_{TA,i}$

    \STATE \textbf{(Error indicator)} $err_i \leftarrow \mathbb{I}\{A(\hat{y}^{(MLG)}_i,B_i^*)=0\}$
\ENDFOR

\STATE Sort uncertainties ascending: $u_{(1)}\le \cdots \le u_{(N)}$ with aligned $err_{(1)},\ldots,err_{(N)}$
% \STATE Initialize $\hat{\tau}\leftarrow$ \texttt{NULL}

% \FOR{$t=1$ \textbf{to} $N$}
%     \STATE Candidate threshold $\tau \leftarrow u_{(t)}$
%     \STATE $n \leftarrow \sum_{j=1}^{N}\mathbb{I}\{u_{(j)}\le \tau\}$ \hfill(\# accepted)
%     \STATE $X \leftarrow \sum_{j=1}^{N}\mathbb{I}\{u_{(j)}\le \tau \land err_{(j)}=1\}$ \hfill(\# errors among accepted)
%     \STATE \textbf{Clopper--Pearson upper bound:} $\mathrm{UCB} \leftarrow \mathrm{BetaInv}(1-\delta;\,X+1,\,n-X)$
%     \IF{$\mathrm{UCB}\le \alpha$}
%         \STATE Update $\hat{\tau}\leftarrow \tau$
%     \ENDIF
% \ENDFOR

% \IF{$\hat{\tau}=$ \texttt{NULL}}
%     \STATE \textbf{Return} ``The target risk level $\alpha$ is unattainable.''
% \ELSE
%     \STATE \textbf{Return} $\hat{\tau}$
% \ENDIF
\STATE Initialize the selected threshold $\hat{\tau} \leftarrow \texttt{NULL}$

\FOR{$t = 1$ \textbf{to} $N$}
    \STATE Set candidate threshold $\tau \leftarrow u_{(t)}$
    \STATE $n \leftarrow \sum_{j=1}^{N} \mathbb{I}\{u_{(j)} \le \tau\}$ \hfill (number of accepted samples)
    \STATE $X \leftarrow \sum_{j=1}^{N} \mathbb{I}\{u_{(j)} \le \tau \land err_{(j)} = 1\}$ \hfill (number of errors among accepted)
    \STATE \textbf{Compute Clopper--Pearson upper bound:} $\mathrm{UCB} \leftarrow \mathrm{BetaInv}(1-\delta;\, X+1,\, n-X)$
    \IF{$\mathrm{UCB} \le \alpha$}
        \STATE Update $\hat{\tau} \leftarrow \tau$ 
    \ENDIF
\ENDFOR

\IF{$\hat{\tau} = \texttt{NULL}$}
    \STATE \textbf{Return} ``The target risk level $\alpha$ is unattainable under calibration.''
\ELSE
    \STATE \textbf{Return} $\hat{\tau}$
\ENDIF

\end{algorithmic}
\end{algorithm}

%% file: section/Appendix/d.tex
\section{Prompt Template}
\begin{figure*}[!t]
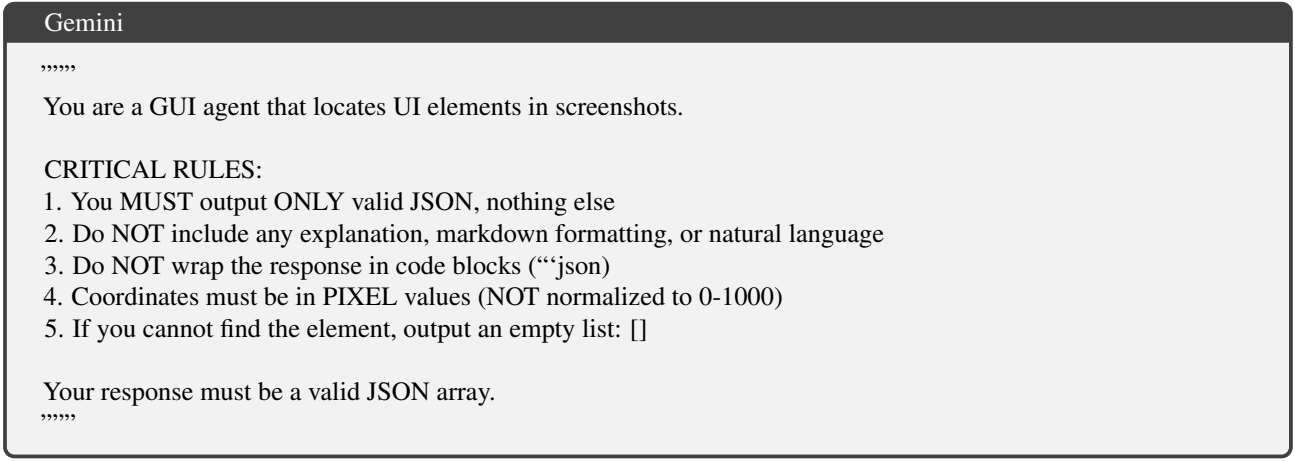

\begin{tcolorbox}[title=Gemini]
"""\\
You are a GUI agent that locates UI elements in screenshots.\\

CRITICAL RULES:\\
1. You MUST output ONLY valid JSON, nothing else\\
2. Do NOT include any explanation, markdown formatting, or natural language\\
3. Do NOT wrap the response in code blocks (```json)\\
4. Coordinates must be in PIXEL values (NOT normalized to 0-1000)\\
5. If you cannot find the element, output an empty list: []\\

Your response must be a valid JSON array.\\
"""
\end{tcolorbox}
\caption{A system prompt example for Gemini-3-pro in ScreenSpot-Pro dataset.}
\label{fig: gemini system prompt}
\end{figure*}

To ensure a fair and reliable evaluation of large vision--language models on GUI grounding, we adopt a strictly constrained prompt template for Gemini in the ScreenSpot-Pro benchmark, as illustrated in Figure~\ref{fig: gemini system prompt},~\ref{fig: gemini user prompt}.

\begin{figure*}[!t]
\begin{tcolorbox}[title=Gemini]
Task: Point to the UI element matching this instruction: \texttt{\{instruction\}} \\

Image size: \texttt{\{W\} x \{H\} pixels.} \\

Output format (JSON only, no markdown):\\
 \texttt{[\{"point": [y, x], "label": "description"\}]} \\

Where: \\
- point: \texttt{[y, x] coordinates in PIXELS (NOT normalized to 0-1000)} \\
- \texttt{y is vertical (0 to \{H\})} \\
- \texttt{x is horizontal (0 to \{W\})} \\

If no element found, output: [] \\

Example: \texttt{[\{"point": [60, 230], "label": "submit button"\}]} \\
\end{tcolorbox}
\caption{A user prompt example for Gemini-3-pro in ScreenSpot-Pro dataset.}
\label{fig: gemini user prompt}
\end{figure*}

%% file: section/Appendix/e.tex
\section{Case Study}
\begin{figure*}[!t]
    \centering
    \includegraphics[width=\textwidth, trim=7cm 7cm 7.5cm 7cm,clip]{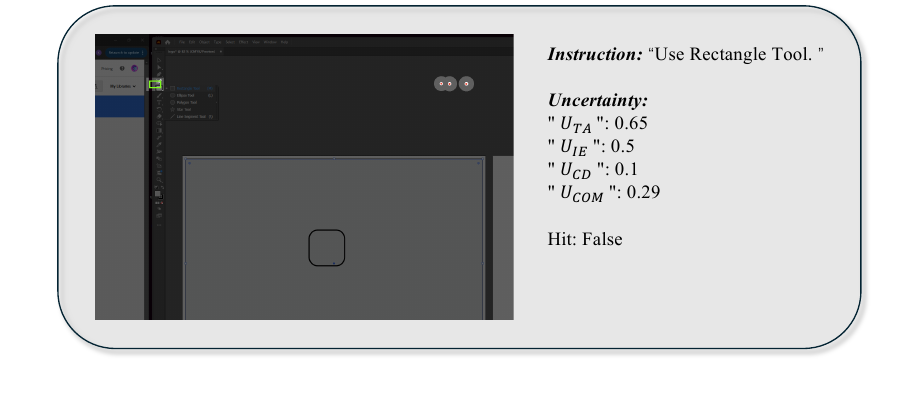}
    \caption{
    An example of GUI grounding task using our uncertainty score.
    }
    \label{fig:ex1}
\end{figure*}

\begin{figure*}[!t]
    \centering
    \includegraphics[width=\textwidth, trim=7.5cm 7cm 7cm 7cm,clip]{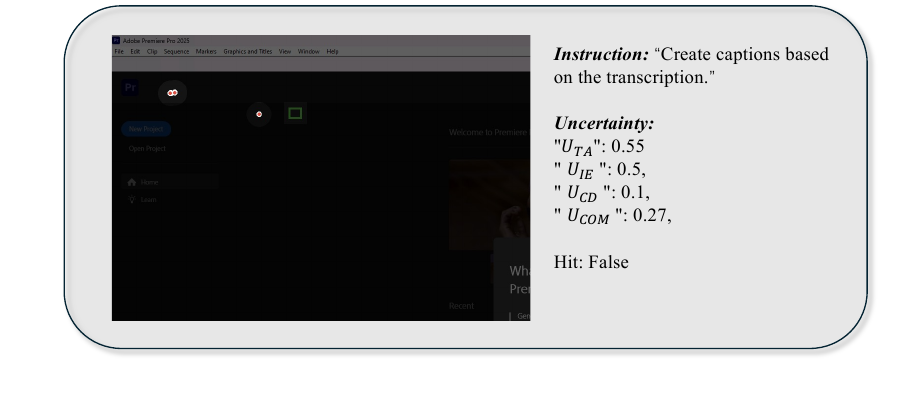}
    \caption{
    An example of GUI grounding task using our uncertainty score.
    }
    \label{fig:ex2}
\end{figure*}

\begin{figure*}[!t]
    \centering
    \includegraphics[width=\textwidth, trim=7.5cm 7cm 7cm 7cm,clip]{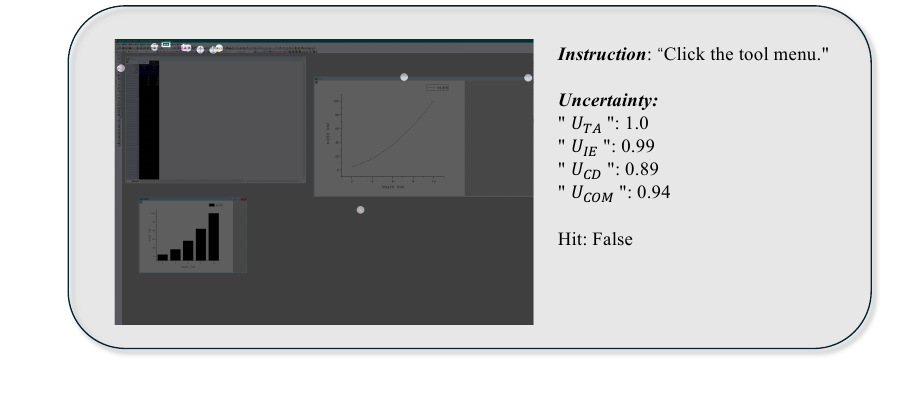}
    \caption{
    An example of GUI grounding task using our uncertainty score.
    }
    \label{fig:ex3}
\end{figure*}

\begin{figure*}[!t]
    \centering
    \includegraphics[width=\textwidth, trim=7.5cm 7.5cm 6.5cm 7cm,clip]{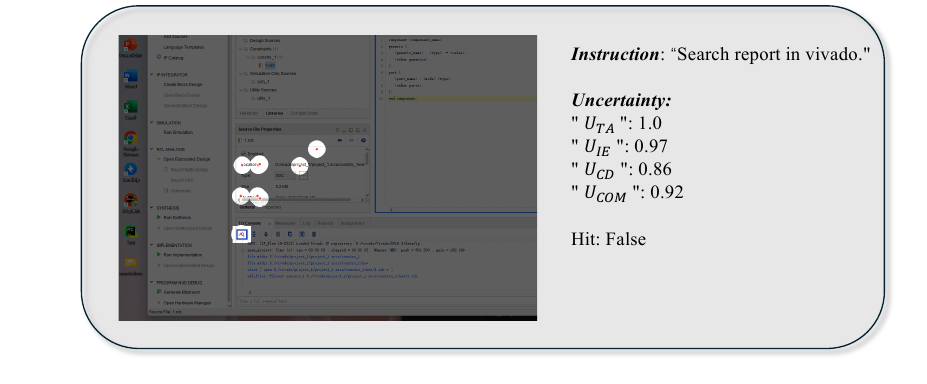}
    \caption{
    An example of GUI grounding task using our uncertainty score.
    }
    \label{fig:ex4}
\end{figure*}

\begin{figure*}[!t]
    \centering
    \includegraphics[width=\textwidth, trim=7cm 7.5cm 7cm 6cm,clip]{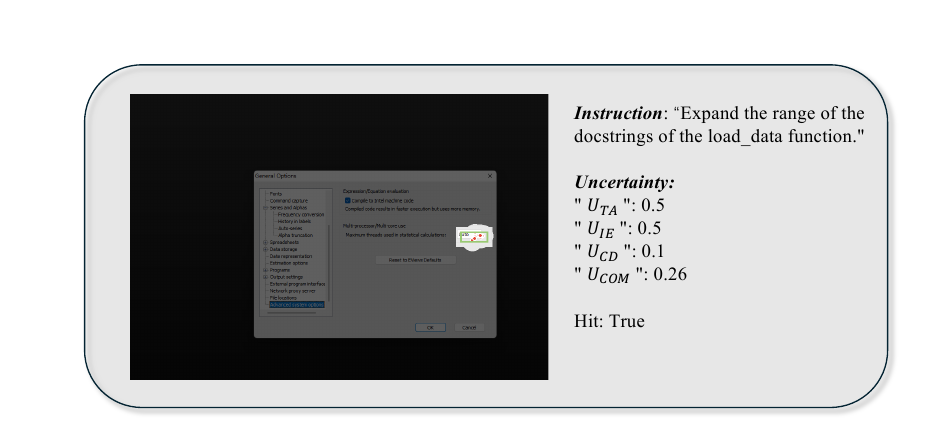}
    \caption{
    An example of GUI grounding task using our uncertainty score.
    }
    \label{fig:ex5}
\end{figure*}

We present qualitative examples to illustrate how the proposed uncertainty score reflects the reliability of GUI grounding predictions in practice in Figure~\ref{fig:ex1},~\ref{fig:ex2},~\ref{fig:ex3},~\ref{fig:ex4},~\ref{fig:ex5}.


%% file: section/Appendix/f.tex
\section{Additional Experimental Results}
\label{app:ex}
% \subsection{Exploratory Study: Uncertainty-Guided Spatial Zoom-In}
% \label{sec:zoom_in}
% In cascaded inference, deferring uncertain cases to a powerful expert model can improve accuracy but incurs non-trivial cost.
% Beyond deciding when to defer, we briefly explore whether uncertainty can also inform where the expert model should focus.

% For each deferred instance, we leverage the sampling-based spatial uncertainty map produced by \textsc{SafeGround}.
% We extract activated regions from the uncertainty heatmap and compute the minimal bounding box enclosing these regions.
% The cropped image is forwarded to the expert model, whose predicted coordinates are then mapped back to the original screen.
% This procedure relies solely on uncertainty-derived spatial information and introduces no additional supervision.

% Empirically, this uncertainty-guided zoom-in strategy yields consistent gains for the Holo1.5 family, where stochastic sampling produces spatially dispersed predictions and compact activated regions.
% % For other models with highly deterministic or already well-localized outputs (e.g., GUI-Actor, UI-TARS), the effect is marginal.

% Given its model-dependent behavior, we treat spatial zoom-in as an optional, heuristic extension rather than a core component of \textsc{SafeGround}, and include it here as an exploratory demonstration of how spatial uncertainty can support more efficient fallback execution.
\paragraph{Sensitivity to Sampling Temperature.}
We further examine the sensitivity of the proposed uncertainty measures to the sampling temperature used during stochastic decoding.
Table~\ref{tab:temp_auroc} and Table~\ref{tab:temp_auarc} report AUROC and AUARC results on Holo1.5-3B~\citep{hai2025holo15modelfamily} under different temperature settings.
As the temperature increases, $U_{IE}$ and $U_{CD}$ become more informative, reflected by consistent gains in AUROC.
In contrast, margin-based uncertainty exhibits relatively limited sensitivity to temperature changes.
$U_{COM}$ shows a dependence on the sampling temperature, reflecting its ability to adapt to changes in the diversity and dispersion of stochastic predictions, while remaining competitive across the evaluated temperature range.

\input{tables/appendix/auroc}
\input{tables/appendix/auarc}

\paragraph{Sensitivity to Uncertainty Weighting.}
\label{weightsensitivity}
We examine the sensitivity of the proposed framework to the weighting scheme used in the combined uncertainty score $U_{\mathrm{COM}}$. Starting from the default setting $(w_{CD}, w_{IE}, w_{TA}) = (0.6, 0.2, 0.2)$, we evaluate several alternative weighting configurations that moderately vary the relative contributions of the three uncertainty components, while keeping the weights normalized.

Specifically, we consider the following weighting configurations for the combined uncertainty score $U_{\mathrm{COM}}$:
\begin{itemize}
    \item \textbf{v1:} $(w_{CD}, w_{IE}, w_{TA}) = (0.34, 0.33, 0.33)$;
    \item \textbf{v2:} $(w_{CD}, w_{IE}, w_{TA}) = (0.2, 0.2, 0.6)$;
    \item \textbf{v3:} $(w_{CD}, w_{IE}, w_{TA}) = (0.2, 0.6, 0.2)$;
    \item \textbf{v4:} $(w_{CD}, w_{IE}, w_{TA}) = (0.5, 0.25, 0.25)$;
    \item \textbf{v5:} $(w_{CD}, w_{IE}, w_{TA}) = (0.25, 0.25, 0.5)$;
    \item \textbf{v6:} $(w_{CD}, w_{IE}, w_{TA}) = (0.25, 0.5, 0.25)$;
    \item \textbf{original:} $(w_{CD}, w_{IE}, w_{TA}) = (0.6, 0.2, 0.2)$.
\end{itemize}

\begin{figure}[!t]
    \centering
    \includegraphics[width=0.65\columnwidth]{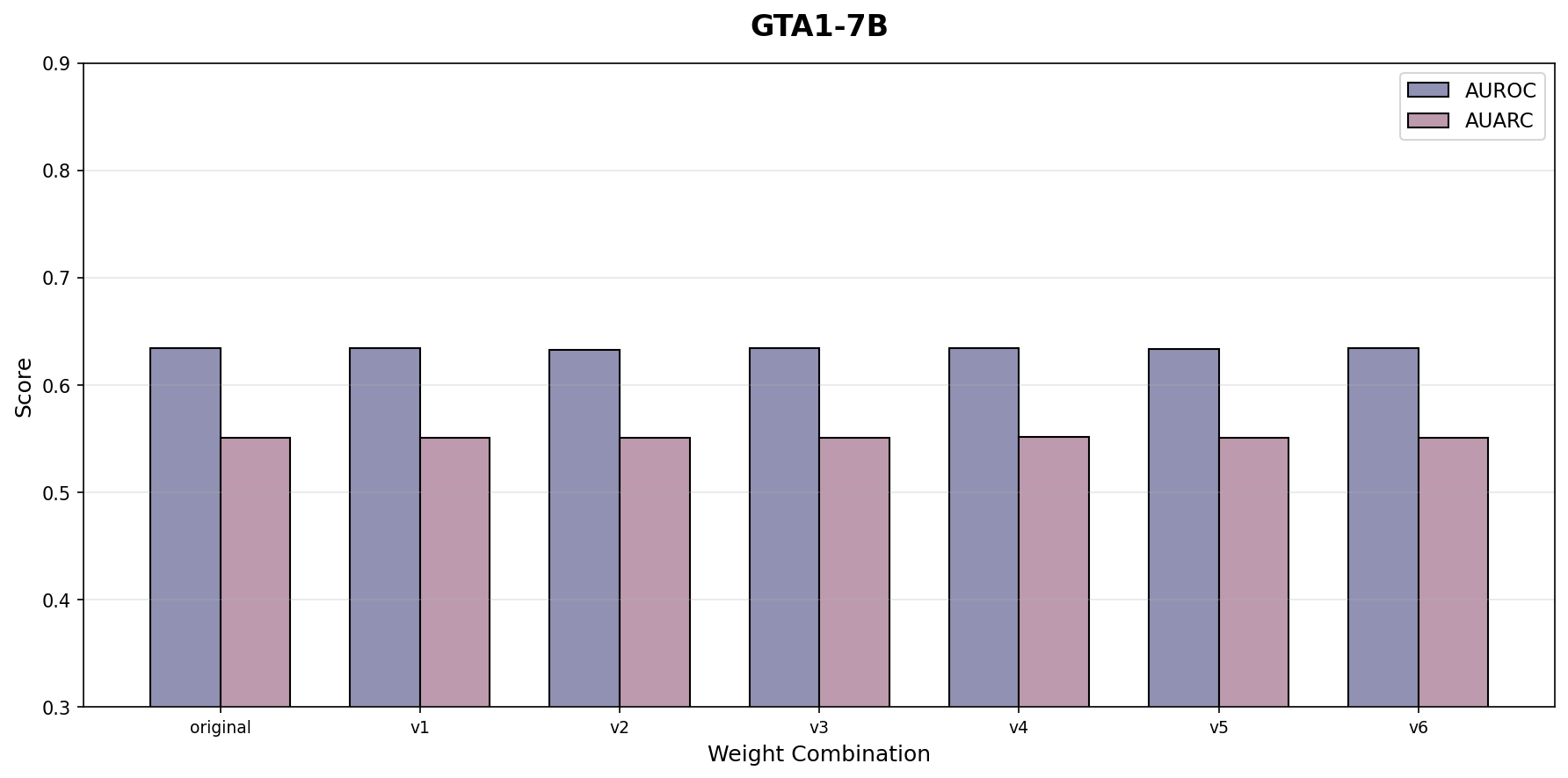}
    \caption{Sensitivity analysis of AUROC and AUARC to uncertainty weighting for GTA1-7B.}
    \label{fig:GTA1}
\end{figure}

\begin{figure}[!t]
    \centering
    \includegraphics[width=0.65\columnwidth]{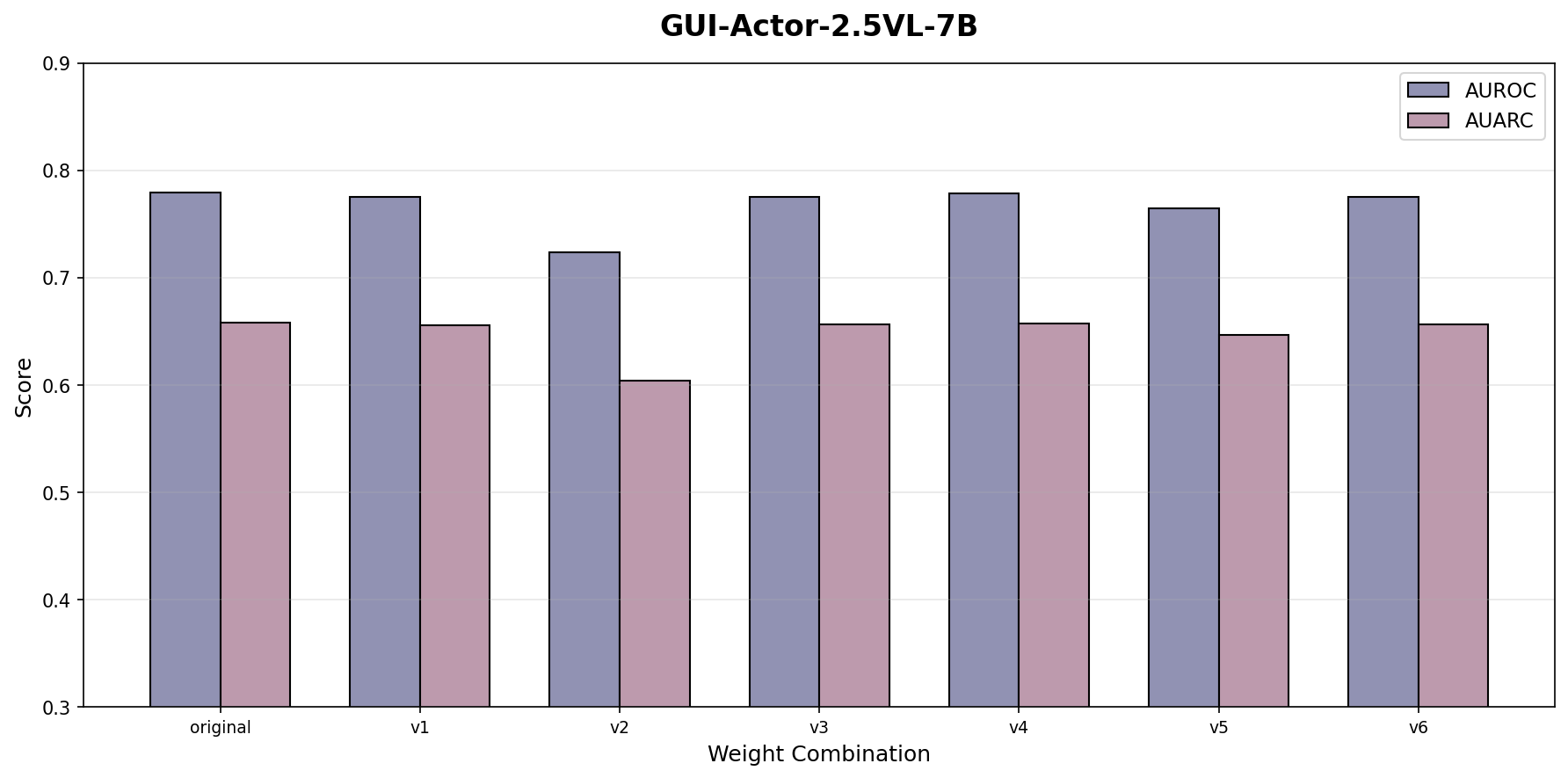}
    \caption{Sensitivity analysis of AUROC and AUARC to uncertainty weighting for GUI-Actor-2.5VL-7B.}
    \label{fig:GUI-Actor-2.5VL}
\end{figure}

\begin{figure}[!t]
    \centering
    \includegraphics[width=0.65\columnwidth]{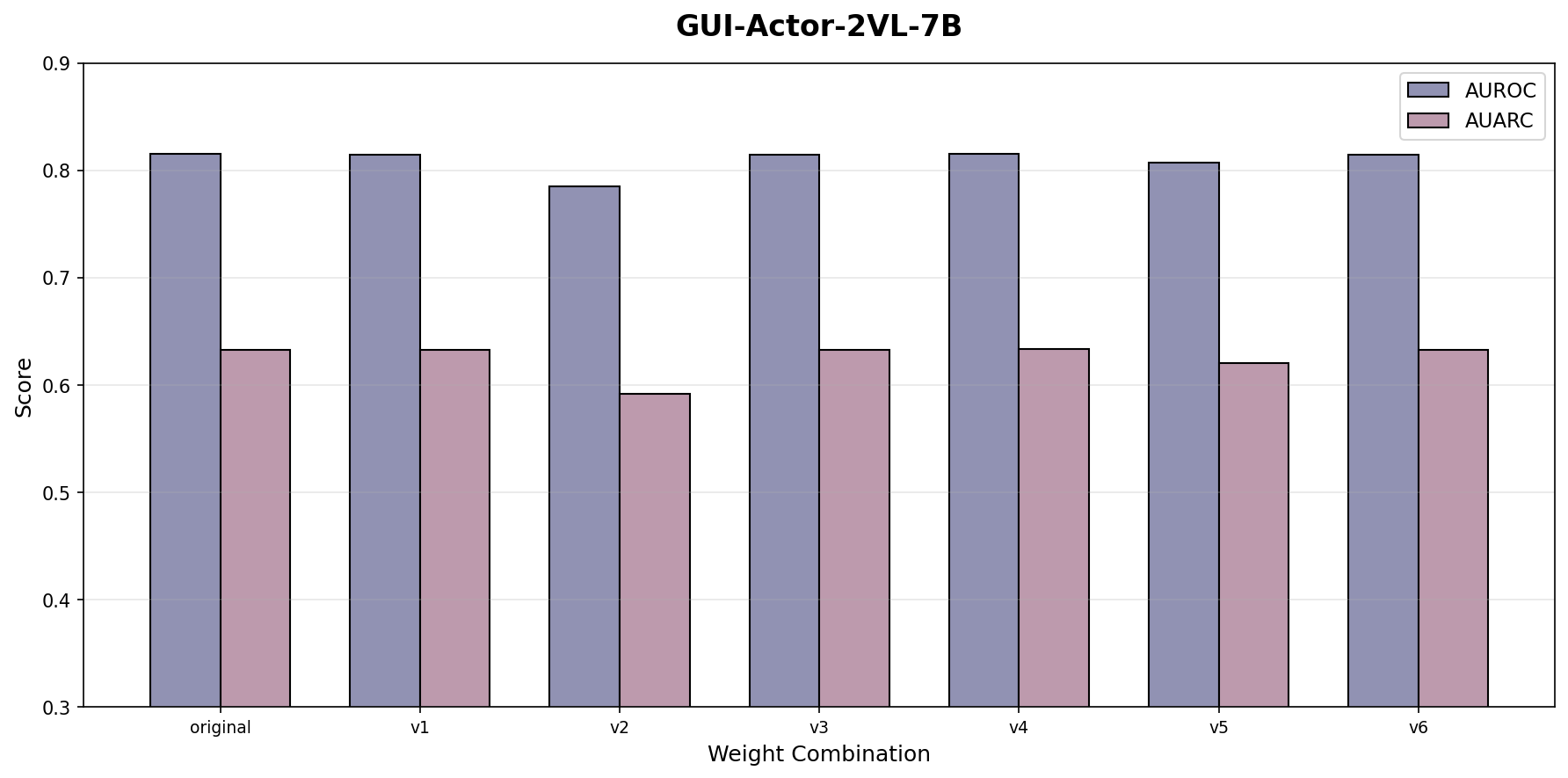}
    \caption{Sensitivity analysis of AUROC and AUARC to uncertainty weighting for GUI-Actor-2VL-7B.}
    \label{fig:GUI-Actor-2VL}
\end{figure}

\begin{figure}[!t]
    \centering
    \includegraphics[width=0.65\columnwidth]{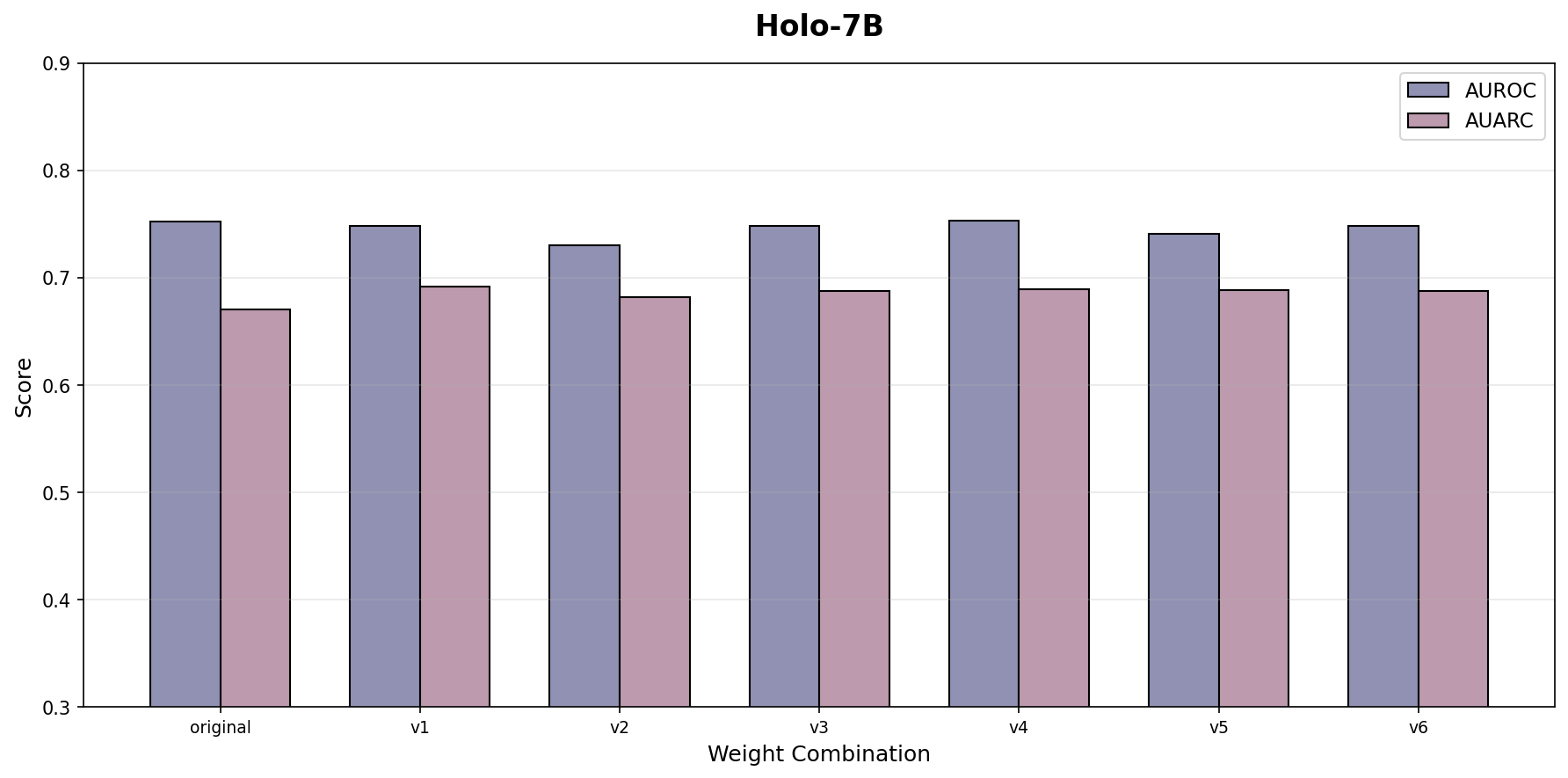}
    \caption{Sensitivity analysis of AUROC and AUARC to uncertainty weighting for Holo1.5-7B.}
    \label{fig:Holo-7B}
\end{figure}

\begin{figure}[!t]
    \centering
    \includegraphics[width=0.65\columnwidth]{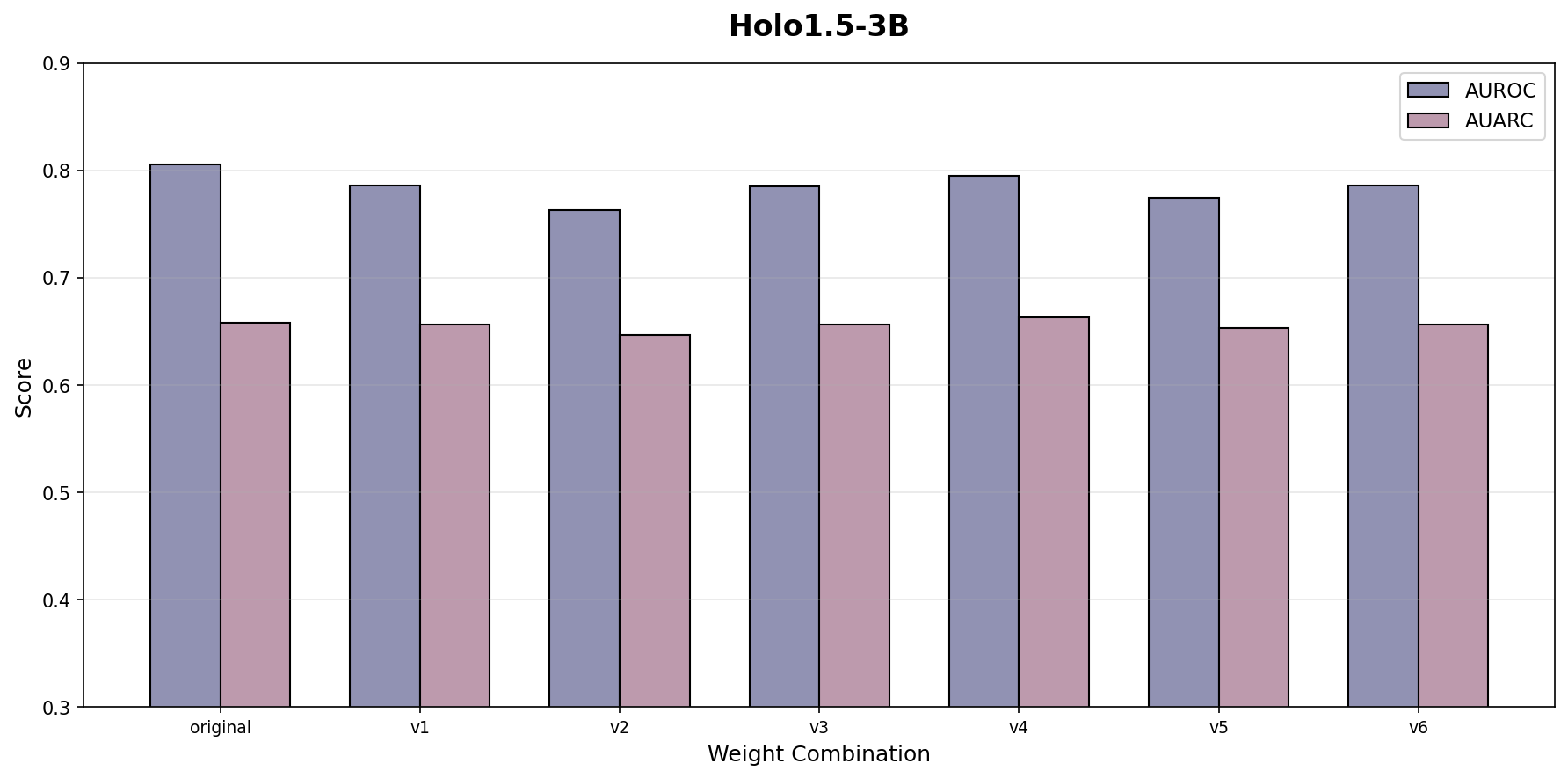}
    \caption{Sensitivity analysis of AUROC and AUARC to uncertainty weighting for Holo1.5-3B.}
    \label{fig:Holo-3B}
\end{figure}

\begin{figure}[!t]
    \centering
    \includegraphics[width=0.65\columnwidth]{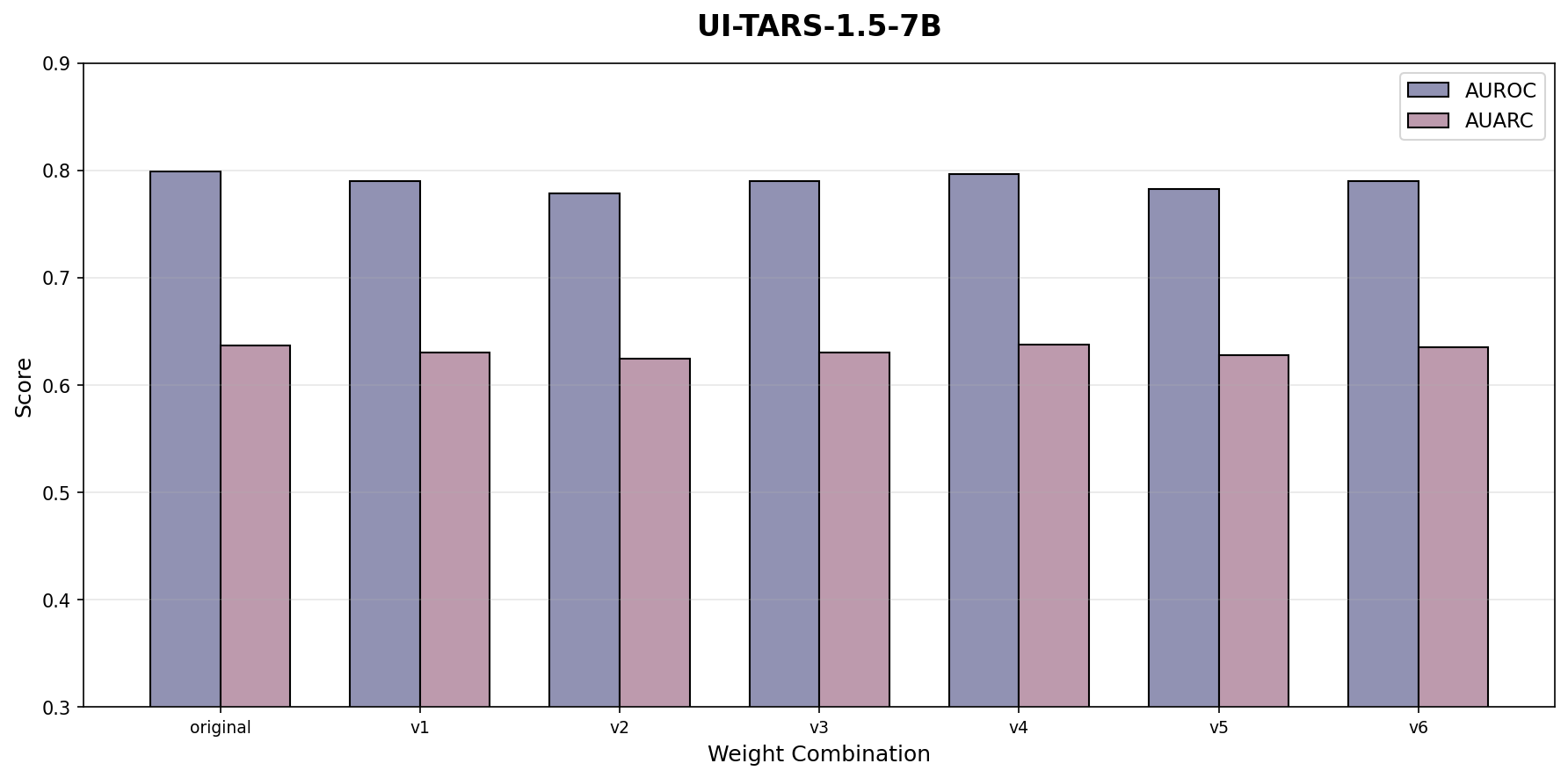}
    \caption{Sensitivity analysis of AUROC and AUARC to uncertainty weighting for UI-TARS-1.5-7B.}
    \label{fig:UI-TARS-1.5}
\end{figure}

As shown in Figure~\ref{fig:GTA1},~\ref{fig:GUI-Actor-2.5VL},~\ref{fig:Holo-7B},~\ref{fig:Holo-3B},~\ref{fig:GUI-Actor-2VL},~\ref{fig:UI-TARS-1.5}, across all evaluated models, both AUROC and AUARC exhibit only minor fluctuations under different weighting schemes. These results indicate that the proposed uncertainty aggregation is robust to moderate changes in the weighting scheme, supporting the use of a fixed, model-agnostic combination in practice.


%% file: tables/appendix/auroc.tex
\begin{table}[t]
\centering
\small
\caption{AUROC of different uncertainty measures on Holo1.5-3B under varying sampling temperatures.}
\label{tab:temp_auroc}
\begin{tabular}{lcccc}
\toprule
\textbf{Method} & \textbf{Temp=0.3} & \textbf{Temp=0.5} & \textbf{Temp=0.7} & \textbf{Temp=1.0} \\
\midrule
$U_{TA}$        & 0.6258 & 0.6270 & 0.6297 & 0.6404 \\
$U_{IE}$       & 0.6621 & 0.6900 & 0.7329 & 0.7753 \\
$U_{CD}$ & 0.6689 & 0.7078 & 0.7590 & 0.8060 \\
$U_{COM}$      & 0.6819 & 0.7218 & 0.7578 & 0.8056 \\
\bottomrule
\end{tabular}
\end{table}

%% file: tables/appendix/auarc.tex
\begin{table}[t]
\centering
\small
\caption{AUARC of different uncertainty measures on Holo1.5-3B under varying sampling temperatures.}
\label{tab:temp_auarc}
\begin{tabular}{lcccc}
\toprule
\textbf{Method} & \textbf{Temp=0.3} & \textbf{Temp=0.5} & \textbf{Temp=0.7} & \textbf{Temp=1.0} \\
\midrule
$U_{TA}$        & 0.5373 & 0.5247 & 0.5186 & 0.5709 \\
$U_{IE}$       & 0.5182 & 0.5165 & 0.5015 & 0.6534 \\
$U_{CD}$       & 0.5219 & 0.5205 & 0.4977 & 0.6578 \\
$U_{COM}$      & 0.5250 & 0.5308 & 0.4960 & 0.6576 \\
\bottomrule
\end{tabular}
\end{table}

%% file: example_paper.bib
@article{wu2025gui,
  title={GUI-Actor: Coordinate-Free Visual Grounding for GUI Agents},
  author={Wu, Qianhui and Cheng, Kanzhi and Yang, Rui and Zhang, Chaoyun and Yang, Jianwei and Jiang, Huiqiang and Mu, Jian and Peng, Baolin and Qiao, Bo and Tan, Reuben and others},
  journal={arXiv preprint arXiv:2506.03143},
  year={2025}
}

@misc{
zhang2025hyperclick,
title={HyperClick: Advancing Reliable {GUI} Grounding via Uncertainty Calibration},
author={Shaojie Zhang and Pei Fu and Ruoceng Zhang and Jiahui Yang and Anan Du and Xiuwen Xi and Shaokang Wang and Ying Huang and Bin Qin and Zhenbo Luo and Jian Luan},
year={2025},
url={https://openreview.net/forum?id=pXYwksqDyE}
}

@misc{chen2023shikraunleashingmultimodalllms,
      title={Shikra: Unleashing Multimodal LLM's Referential Dialogue Magic}, 
      author={Keqin Chen and Zhao Zhang and Weili Zeng and Richong Zhang and Feng Zhu and Rui Zhao},
      year={2023},
      eprint={2306.15195},
      archivePrefix={arXiv},
      primaryClass={cs.CV},
      url={https://arxiv.org/abs/2306.15195}, 
}

@misc{wang2024qwen2vlenhancingvisionlanguagemodels,
      title={Qwen2-VL: Enhancing Vision-Language Model's Perception of the World at Any Resolution}, 
      author={Peng Wang and Shuai Bai and Sinan Tan and Shijie Wang and Zhihao Fan and Jinze Bai and Keqin Chen and Xuejing Liu and Jialin Wang and Wenbin Ge and Yang Fan and Kai Dang and Mengfei Du and Xuancheng Ren and Rui Men and Dayiheng Liu and Chang Zhou and Jingren Zhou and Junyang Lin},
      year={2024},
      eprint={2409.12191},
      archivePrefix={arXiv},
      primaryClass={cs.CV},
      url={https://arxiv.org/abs/2409.12191}, 
}

@misc{qin2025uitarspioneeringautomatedgui,
      title={UI-TARS: Pioneering Automated GUI Interaction with Native Agents}, 
      author={Yujia Qin and Yining Ye and Junjie Fang and Haoming Wang and Shihao Liang and Shizuo Tian and Junda Zhang and Jiahao Li and Yunxin Li and Shijue Huang and Wanjun Zhong and Kuanye Li and Jiale Yang and Yu Miao and Woyu Lin and Longxiang Liu and Xu Jiang and Qianli Ma and Jingyu Li and Xiaojun Xiao and Kai Cai and Chuang Li and Yaowei Zheng and Chaolin Jin and Chen Li and Xiao Zhou and Minchao Wang and Haoli Chen and Zhaojian Li and Haihua Yang and Haifeng Liu and Feng Lin and Tao Peng and Xin Liu and Guang Shi},
      year={2025},
      eprint={2501.12326},
      archivePrefix={arXiv},
      primaryClass={cs.AI},
      url={https://arxiv.org/abs/2501.12326}, 
}

@misc{nguyen2025guiagentssurvey,
      title={GUI Agents: A Survey}, 
      author={Dang Nguyen and Jian Chen and Yu Wang and Gang Wu and Namyong Park and Zhengmian Hu and Hanjia Lyu and Junda Wu and Ryan Aponte and Yu Xia and Xintong Li and Jing Shi and Hongjie Chen and Viet Dac Lai and Zhouhang Xie and Sungchul Kim and Ruiyi Zhang and Tong Yu and Mehrab Tanjim and Nesreen K. Ahmed and Puneet Mathur and Seunghyun Yoon and Lina Yao and Branislav Kveton and Jihyung Kil and Thien Huu Nguyen and Trung Bui and Tianyi Zhou and Ryan A. Rossi and Franck Dernoncourt},
      year={2025},
      eprint={2412.13501},
      archivePrefix={arXiv},
      primaryClass={cs.AI},
      url={https://arxiv.org/abs/2412.13501}, 
}

@misc{liu2025uncertaintyquantificationconfidencecalibration,
      title={Uncertainty Quantification and Confidence Calibration in Large Language Models: A Survey}, 
      author={Xiaoou Liu and Tiejin Chen and Longchao Da and Chacha Chen and Zhen Lin and Hua Wei},
      year={2025},
      eprint={2503.15850},
      archivePrefix={arXiv},
      primaryClass={cs.CL},
      url={https://arxiv.org/abs/2503.15850}, 
}

@article{clopper1934use,
  title={The use of confidence or fiducial limits illustrated in the case of the binomial},
  author={Clopper, Charles J and Pearson, Egon S},
  journal={Biometrika},
  year={1934}
}

@misc{angelopoulos2022learntestcalibratingpredictive,
      title={Learn then Test: Calibrating Predictive Algorithms to Achieve Risk Control}, 
      author={Anastasios N. Angelopoulos and Stephen Bates and Emmanuel J. Candès and Michael I. Jordan and Lihua Lei},
      year={2022},
      eprint={2110.01052},
      archivePrefix={arXiv},
      primaryClass={cs.LG},
      url={https://arxiv.org/abs/2110.01052}, 
}

@misc{wang2025saferriskconstrainedsamplethenfilterlarge,
      title={SAFER: Risk-Constrained Sample-then-Filter in Large Language Models}, 
      author={Qingni Wang and Yue Fan and Xin Eric Wang},
      year={2025},
      eprint={2510.10193},
      archivePrefix={arXiv},
      primaryClass={cs.AI},
      url={https://arxiv.org/abs/2510.10193}, 
}

@article{10.1016/j.ijhcs.2025.103455,
author = {Xu, Zhengtao and Song, Tianqi and Lee, Yi-Chieh},
title = {Confronting verbalized uncertainty: Understanding how LLM’s verbalized uncertainty influences users in AI-assisted decision-making},
year = {2025},
issue_date = {Mar 2025},
publisher = {Academic Press, Inc.},
address = {USA},
volume = {197},
number = {C},
issn = {1071-5819},
url = {https://doi.org/10.1016/j.ijhcs.2025.103455},
doi = {10.1016/j.ijhcs.2025.103455},
journal = {Int. J. Hum.-Comput. Stud.},
month = mar,
numpages = {17}
}

@misc{kuhn2023semanticuncertaintylinguisticinvariances,
      title={Semantic Uncertainty: Linguistic Invariances for Uncertainty Estimation in Natural Language Generation}, 
      author={Lorenz Kuhn and Yarin Gal and Sebastian Farquhar},
      year={2023},
      eprint={2302.09664},
      archivePrefix={arXiv},
      primaryClass={cs.CL},
      url={https://arxiv.org/abs/2302.09664}, 
}

@misc{hou2025probabilisticframeworkllmhallucination,
      title={A Probabilistic Framework for LLM Hallucination Detection via Belief Tree Propagation}, 
      author={Bairu Hou and Yang Zhang and Jacob Andreas and Shiyu Chang},
      year={2025},
      eprint={2406.06950},
      archivePrefix={arXiv},
      primaryClass={cs.CL},
      url={https://arxiv.org/abs/2406.06950}, 
}

@misc{wang2024wordsequenceentropyuncertaintyestimation,
      title={Word-Sequence Entropy: Towards Uncertainty Estimation in Free-Form Medical Question Answering Applications and Beyond}, 
      author={Zhiyuan Wang and Jinhao Duan and Chenxi Yuan and Qingyu Chen and Tianlong Chen and Yue Zhang and Ren Wang and Xiaoshuang Shi and Kaidi Xu},
      year={2024},
      eprint={2402.14259},
      archivePrefix={arXiv},
      primaryClass={cs.CL},
      url={https://arxiv.org/abs/2402.14259}, 
}

@misc{angelopoulos2022gentleintroductionconformalprediction,
      title={A Gentle Introduction to Conformal Prediction and Distribution-Free Uncertainty Quantification}, 
      author={Anastasios N. Angelopoulos and Stephen Bates},
      year={2022},
      eprint={2107.07511},
      archivePrefix={arXiv},
      primaryClass={cs.LG},
      url={https://arxiv.org/abs/2107.07511}, 
}

@inproceedings{
jung2025trust,
title={Trust or Escalate: {LLM} Judges with Provable Guarantees for Human Agreement},
author={Jaehun Jung and Faeze Brahman and Yejin Choi},
booktitle={The Thirteenth International Conference on Learning Representations},
year={2025},
url={https://openreview.net/forum?id=UHPnqSTBPO}
}

@misc{wang2025coinuncertaintyguardingselectivequestion,
      title={COIN: Uncertainty-Guarding Selective Question Answering for Foundation Models with Provable Risk Guarantees}, 
      author={Zhiyuan Wang and Jinhao Duan and Qingni Wang and Xiaofeng Zhu and Tianlong Chen and Xiaoshuang Shi and Kaidi Xu},
      year={2025},
      eprint={2506.20178},
      archivePrefix={arXiv},
      primaryClass={cs.CL},
      url={https://arxiv.org/abs/2506.20178}, 
}

@inproceedings{fan-etal-2025-gui,
    title = "{GUI}-Bee: Align {GUI} Action Grounding to Novel Environments via Autonomous Exploration",
    author = "Fan, Yue  and
      Zhao, Handong  and
      Zhang, Ruiyi  and
      Shen, Yu  and
      Wang, Xin Eric  and
      Wu, Gang",
    editor = "Christodoulopoulos, Christos  and
      Chakraborty, Tanmoy  and
      Rose, Carolyn  and
      Peng, Violet",
    booktitle = "Proceedings of the 2025 Conference on Empirical Methods in Natural Language Processing",
    month = nov,
    year = "2025",
    address = "Suzhou, China",
    publisher = "Association for Computational Linguistics"
}

@InProceedings{pmlr-v48-gal16,
  title = 	 {Dropout as a Bayesian Approximation: Representing Model Uncertainty in Deep Learning},
  author = 	 {Gal, Yarin and Ghahramani, Zoubin},
  booktitle = 	 {Proceedings of The 33rd International Conference on Machine Learning}
}

@inproceedings{
wang2025sample,
title={Sample then Identify: A General Framework for Risk Control and Assessment in Multimodal Large Language Models},
author={Qingni Wang and Tiantian Geng and Zhiyuan Wang and Teng Wang and Bo Fu and Feng Zheng},
booktitle={The Thirteenth International Conference on Learning Representations},
year={2025},
url={https://openreview.net/forum?id=9WYMDgxDac}
}

@misc{hai2025holo15modelfamily,
      title={Holo1.5 - Open Foundation Models for Computer Use Agents}, 
      author={H Company},
      year={2025},
      url={https://huggingface.co/collections/Hcompany/holo15-68c1a5736e8583a309d23d9b}, 
}

@misc{wu2025guiactorcoordinatefreevisualgrounding,
      title={GUI-Actor: Coordinate-Free Visual Grounding for GUI Agents}, 
      author={Qianhui Wu and Kanzhi Cheng and Rui Yang and Chaoyun Zhang and Jianwei Yang and Huiqiang Jiang and Jian Mu and Baolin Peng and Bo Qiao and Reuben Tan and Si Qin and Lars Liden and Qingwei Lin and Huan Zhang and Tong Zhang and Jianbing Zhang and Dongmei Zhang and Jianfeng Gao},
      year={2025},
      eprint={2506.03143},
      archivePrefix={arXiv},
      primaryClass={cs.CL},
      url={https://arxiv.org/abs/2506.03143}, 
}

@article{qin2025ui,
  title={UI-TARS: Pioneering Automated GUI Interaction with Native Agents},
  author={Qin, Yujia and Ye, Yining and Fang, Junjie and Wang, Haoming and Liang, Shihao and Tian, Shizuo and Zhang, Junda and Li, Jiahao and Li, Yunxin and Huang, Shijue and others},
  journal={arXiv preprint arXiv:2501.12326},
  year={2025}
}

@misc{yang2025gta1guitesttimescaling,
      title={GTA1: GUI Test-time Scaling Agent}, 
      author={Yan Yang and Dongxu Li and Yutong Dai and Yuhao Yang and Ziyang Luo and Zirui Zhao and Zhiyuan Hu and Junzhe Huang and Amrita Saha and Zeyuan Chen and Ran Xu and Liyuan Pan and Silvio Savarese and Caiming Xiong and Junnan Li},
      year={2025},
      eprint={2507.05791},
      archivePrefix={arXiv},
      primaryClass={cs.AI},
      url={https://arxiv.org/abs/2507.05791}, 
}

@misc{li2025screenspotproguigroundingprofessional,
      title={ScreenSpot-Pro: GUI Grounding for Professional High-Resolution Computer Use}, 
      author={Kaixin Li and Ziyang Meng and Hongzhan Lin and Ziyang Luo and Yuchen Tian and Jing Ma and Zhiyong Huang and Tat-Seng Chua},
      year={2025},
      eprint={2504.07981},
      archivePrefix={arXiv},
      primaryClass={cs.CV},
      url={https://arxiv.org/abs/2504.07981}, 
}

@misc{lin2024generatingconfidenceuncertaintyquantification,
      title={Generating with Confidence: Uncertainty Quantification for Black-box Large Language Models}, 
      author={Zhen Lin and Shubhendu Trivedi and Jimeng Sun},
      year={2024},
      eprint={2305.19187},
      archivePrefix={arXiv},
      primaryClass={cs.CL},
      url={https://arxiv.org/abs/2305.19187}, 
}

@article{team2023gemini,
  title={Gemini: a family of highly capable multimodal models},
  author={Team, Gemini and Anil, Rohan and Borgeaud, Sebastian and Alayrac, Jean-Baptiste and Yu, Jiahui and Soricut, Radu and Schalkwyk, Johan and Dai, Andrew M and Hauth, Anja and Millican, Katie and others},
  journal={arXiv preprint arXiv:2312.11805},
  year={2023}
}

@inproceedings{cheng2024seeclick,
  title={Seeclick: Harnessing gui grounding for advanced visual gui agents},
  author={Cheng, Kanzhi and Sun, Qiushi and Chu, Yougang and Xu, Fangzhi and YanTao, Li and Zhang, Jianbing and Wu, Zhiyong},
  booktitle={Proceedings of the 62nd Annual Meeting of the Association for Computational Linguistics (Volume 1: Long Papers)},
  pages={9313--9332},
  year={2024}
}

@inproceedings{wang-etal-2025-sconu,
    title = "{SC}on{U}: Selective Conformal Uncertainty in Large Language Models",
    author = "Wang, Zhiyuan  and
      Wang, Qingni  and
      Zhang, Yue  and
      Chen, Tianlong  and
      Zhu, Xiaofeng  and
      Shi, Xiaoshuang  and
      Xu, Kaidi",
    editor = "Che, Wanxiang  and
      Nabende, Joyce  and
      Shutova, Ekaterina  and
      Pilehvar, Mohammad Taher",
    booktitle = "Proceedings of the 63rd Annual Meeting of the Association for Computational Linguistics (Volume 1: Long Papers)",
    month = jul

}

@InProceedings{Hong_2024_CVPR,
    author    = {Hong, Wenyi and Wang, Weihan and Lv, Qingsong and Xu, Jiazheng and Yu, Wenmeng and Ji, Junhui and Wang, Yan and Wang, Zihan and Dong, Yuxiao and Ding, Ming and Tang, Jie},
    title     = {CogAgent: A Visual Language Model for GUI Agents},
    booktitle = {Proceedings of the IEEE/CVF Conference on Computer Vision and Pattern Recognition (CVPR)},
    month     = {June},
    year      = {2024},
    pages     = {14281-14290}
}

@misc{gawlikowski2022surveyuncertaintydeepneural,
      title={A Survey of Uncertainty in Deep Neural Networks}, 
      author={Jakob Gawlikowski and Cedrique Rovile Njieutcheu Tassi and Mohsin Ali and Jongseok Lee and Matthias Humt and Jianxiang Feng and Anna Kruspe and Rudolph Triebel and Peter Jung and Ribana Roscher and Muhammad Shahzad and Wen Yang and Richard Bamler and Xiao Xiang Zhu},
      year={2022},
      eprint={2107.03342},
      archivePrefix={arXiv},
      primaryClass={cs.LG},
      url={https://arxiv.org/abs/2107.03342}, 
}

@misc{hu2023uncertaintynaturallanguageprocessing,
      title={Uncertainty in Natural Language Processing: Sources, Quantification, and Applications}, 
      author={Mengting Hu and Zhen Zhang and Shiwan Zhao and Minlie Huang and Bingzhe Wu},
      year={2023},
      eprint={2306.04459},
      archivePrefix={arXiv},
      primaryClass={cs.CL},
      url={https://arxiv.org/abs/2306.04459}, 
}

@inproceedings{
ye2024benchmarking,
title={Benchmarking {LLM}s via Uncertainty Quantification},
author={Fanghua Ye and Mingming Yang and Jianhui Pang and Longyue Wang and Derek F. Wong and Emine Yilmaz and Shuming Shi and Zhaopeng Tu},
booktitle={The Thirty-eight Conference on Neural Information Processing Systems Datasets and Benchmarks Track},
year={2024},
url={https://openreview.net/forum?id=L0oSfTroNE}
}

@misc{kadavath2022languagemodelsmostlyknow,
      title={Language Models (Mostly) Know What They Know}, 
      author={Saurav Kadavath and Tom Conerly and Amanda Askell and Tom Henighan and Dawn Drain and Ethan Perez and Nicholas Schiefer and Zac Hatfield-Dodds and Nova DasSarma and Eli Tran-Johnson and Scott Johnston and Sheer El-Showk and Andy Jones and Nelson Elhage and Tristan Hume and Anna Chen and Yuntao Bai and Sam Bowman and Stanislav Fort and Deep Ganguli and Danny Hernandez and Josh Jacobson and Jackson Kernion and Shauna Kravec and Liane Lovitt and Kamal Ndousse and Catherine Olsson and Sam Ringer and Dario Amodei and Tom Brown and Jack Clark and Nicholas Joseph and Ben Mann and Sam McCandlish and Chris Olah and Jared Kaplan},
      year={2022},
      eprint={2207.05221},
      archivePrefix={arXiv},
      primaryClass={cs.CL},
      url={https://arxiv.org/abs/2207.05221}, 
}

@inproceedings{
hendrycks2017a,
title={A Baseline for Detecting Misclassified and Out-of-Distribution Examples in Neural Networks},
author={Dan Hendrycks and Kevin Gimpel},
booktitle={International Conference on Learning Representations},
year={2017},
url={https://openreview.net/forum?id=Hkg4TI9xl}
}

@article{geifman2017selective,
  title={Selective classification for deep neural networks},
  author={Geifman, Yonatan and El-Yaniv, Ran},
  journal={Advances in neural information processing systems},
  volume={30},
  year={2017}
}

@article{lin2023generating,
  title={Generating with confidence: Uncertainty quantification for black-box large language models},
  author={Lin, Zhen and Trivedi, Shubhendu and Sun, Jimeng},
  journal={arXiv preprint arXiv:2305.19187},
  year={2023}
}

@inproceedings{
kuhn2023semantic,
title={Semantic Uncertainty: Linguistic Invariances for Uncertainty Estimation in Natural Language Generation},
author={Lorenz Kuhn and Yarin Gal and Sebastian Farquhar},
booktitle={The Eleventh International Conference on Learning Representations },
year={2023},
url={https://openreview.net/forum?id=VD-AYtP0dve}
}

@misc{band2022benchmarkingbayesiandeeplearning,
      title={Benchmarking Bayesian Deep Learning on Diabetic Retinopathy Detection Tasks}, 
      author={Neil Band and Tim G. J. Rudner and Qixuan Feng and Angelos Filos and Zachary Nado and Michael W. Dusenberry and Ghassen Jerfel and Dustin Tran and Yarin Gal},
      year={2022},
      eprint={2211.12717},
      archivePrefix={arXiv},
      primaryClass={stat.ML},
      url={https://arxiv.org/abs/2211.12717}, 
}

@article{pouget2016confidence,
  title={Confidence and certainty: distinct probabilistic quantities for different goals},
  author={Pouget, Alexandre and Drugowitsch, Jan and Kepecs, Adam},
  journal={Nature neuroscience},
  volume={19},
  number={3},
  pages={366--374},
  year={2016},
  publisher={Nature Publishing Group US New York}
}
